\documentclass[10pt,journal,compsoc]{IEEEtran}

\usepackage{epsfig}
\usepackage{graphicx}
\usepackage{amsmath}
\usepackage{amssymb}
\usepackage{pifont}%
\usepackage{mathrsfs}
\usepackage{color}
\usepackage[colorlinks,linkcolor=black,citecolor=black]{hyperref}
\usepackage{graphics} 
\usepackage{longtable}
\usepackage{rotating}
\usepackage{array}
\usepackage{float}
\usepackage{subfigure}
\usepackage{caption}
\usepackage{xcolor}
\usepackage{indentfirst}
\usepackage{setspace}
\usepackage{algorithmic}

\usepackage{booktabs}
\usepackage{bbding}
\usepackage{multirow}
\usepackage{colortbl}

\usepackage{caption}
\usepackage{wrapfig}
\usepackage[ruled,linesnumbered]{algorithm2e}

\ifCLASSOPTIONcompsoc
  \usepackage[nocompress]{cite}
\else
  \usepackage{cite}
\fi

\ifCLASSINFOpdf
\else
\fi

\begin{document}

\title{A Curriculum-style Self-training Approach for Source-Free Semantic Segmentation}

\author{Yuxi Wang~\IEEEmembership{Member,~IEEE},
    Jian Liang,~\IEEEmembership{Member,~IEEE},
    and Zhaoxiang Zhang,~\IEEEmembership{Senior Member,~IEEE}
\IEEEcompsocitemizethanks{

\IEEEcompsocthanksitem Y. Wang is with the Centre for Artificial Intelligence and Robotics, Hong Kong Institute of Science $\&$ Innovation, Chinese Academy of Sciences (CAIR-HKISI-CAS), Hong Kong 999077, China. Email: yuxiwang93@gmail.com.

\IEEEcompsocthanksitem J. Liang is with the New Laboratory of Pattern Recognition (NLPR), State Key Laboratory of Multimodal Artificial Intelligence Systems, Institute of Automation, Chinese Academy of Sciences (CASIA), Beijing 100190, China, and the University of Chinese Academy of Sciences (UCAS), Beijing 100190, China. Email: liangjian92@gmail.com.

\IEEEcompsocthanksitem Z. Zhang is with the New Laboratory of Pattern Recognition (NLPR), State Key Laboratory of Multimodal Artificial Intelligence Systems, Institute of Automation, Chinese Academy of Sciences (CASIA), Beijing 100190, China, the University of Chinese Academy of Sciences (UCAS), Beijing 100049, China, and the Centre for Artificial Intelligence and Robotics, Hong Kong Institute of Science $\&$ Innovation, Chinese Academy of Sciences (CAIR-HKISI-CAS), Hong Kong 999077, China. Email: zhaoxiang.zhang@ia.ac.cn.

\IEEEcompsocthanksitem Zhaoxiang. Zhang is the corresponding author.}
}

\markboth{IEEE TRANSACTIONS ON PATTERN ANALYSIS AND MACHINE INTELLIGENCE,~Vol.~x, No.~x, July~2024}%
{Shell \MakeLowercase{\textit{et al.}}: Bare Demo of IEEEtran.cls for Computer Society Journals}

\IEEEtitleabstractindextext{%
\begin{abstract}
    Source-free domain adaptation has developed rapidly in recent years, where the well-trained source model is adapted to the target domain instead of the source data, offering the potential for privacy concerns and intellectual property protection. 
    However, a number of feature alignment techniques in prior domain adaptation methods are not feasible in this challenging problem setting.
    Thereby, we resort to probing inherent domain-invariant feature learning and propose a curriculum-style self-training approach for source-free domain adaptive semantic segmentation.
    In particular, we introduce a curriculum-style entropy minimization method to explore the implicit knowledge from the source model, which fits the trained source model to the target data using certain information from easy-to-hard predictions. 
    We then train the segmentation network by the proposed complementary curriculum-style self-training, which utilizes the negative and positive pseudo labels following the curriculum-learning manner. Although negative pseudo-labels with high uncertainty cannot be identified with the correct labels, they can definitely indicate absent classes. 
    Moreover, we employ an information propagation scheme to further reduce the intra-domain discrepancy within the target domain, which could act as a standard post-processing method for the domain adaptation field. 
    Furthermore, we extend the proposed method to a more challenging black-box source model scenario where only the source model's predictions are available. 
    Extensive experiments validate that our method yields state-of-the-art performance on source-free semantic segmentation tasks for both synthetic-to-real and adverse conditions datasets. The code and corresponding trained models are released at \url{https://github.com/yxiwang/ATP}.
\end{abstract}

}

\maketitle

\IEEEdisplaynontitleabstractindextext

\IEEEpeerreviewmaketitle

\IEEEraisesectionheading{\section{Introduction}\label{sec:introduction}}

    \IEEEPARstart{S}{emantic} segmentation is a fundamental task in the computer vision field that aims to estimate the pixel-level predictions for a given image. Despite the rapid progress on deep learning-based methods, attaining high performance usually demands vast amounts of training data with pixel-level annotations. However, collecting large-scale datasets tends to be prohibitively expensive and time-consuming. Alternatively, previous cross-domain learning approaches \cite{FCN,AdaptSeg,CLAN,ADVENT,FDA,CAG,BiMAL,wang2023survey,li2023coarse,gan2022interact} attempt to train a satisfactory segmentation model for unlabeled real-world data (called target domain) by exploiting labeled photo-realistic synthetic images (called source domain).
    However, due to the domain discrepancy, a well-performing model trained on the source domain degrades drastically when it is applied to the target domain. To tackle this issue, various domain alignment strategies are proposed in the domain adaptive semantic segmentation field \cite{hong2018conditional,AdaptSeg,MaxSquare,CyCADA,Liang_2019_CVPR,sakaridis2020map,PyCDA}.
    
    The essential idea for domain adaptive semantic segmentation is to effectively transfer knowledge from a labeled source domain to a distinct unlabeled target domain. Previous methods achieve this goal by bridging the domain gap in the image level \cite{CrDoCo,TGCF-DA,CyCADA,FDA,TextureDA}, feature level \cite{chang2019all,hong2018conditional,MaxSquare}, and output level \cite{AdaptSeg,CLAN,ADVENT,PatchDA}. Despite remarkable progress, they usually require access to the labeled source data during the alignment process. However, in some crucial scenarios, the source data is inaccessible due to data protection laws. For instance, the source datasets of autonomous driving and face recognition are unavailable due to data privacy. Thus, this paper addresses a challenging and interesting source data-free domain adaptation issue, where only the trained source model is provided to the target domain for adaptation instead of the source data.

    Generally, source data-free domain adaptation is more challenging than conventional unsupervised domain adaptation. 
    For the conventional domain adaptation, researchers can reduce domain discrepancy via distribution alignment or adversarial learning method since both source data and target data are available. While in the source data-free scenario, we can only align the target data to the implicit source distribution. 
    Due to the absence of source data, previous feature-level alignment or pixel-level alignment methods are not feasible anymore. Moreover, pseudo labels generated from the source-only model are unreliable because of the performance degradation, so vanilla self-training approaches tend to be harmed by the noisy pseudo labels. An intuitive way to tackle this problem is to utilize the easy transfer samples, enhancing the target model capability first and then propagating the information to the hard transfer samples. To be specific, we first use easy samples to encourage confident predictions and then exploit these reliable predictions to fit the model to the whole target data. We also extend our approach to the black-box source model scenario, where only the predictions of the source model are available. This problem becomes more challenging but more flexible and practical in the rule of the server-and-client paradigm.
    
    In this paper, we propose a curriculum-style self-training framework called \textbf{ATP} to tackle source data-free domain adaptive semantic segmentation, which focuses on enhancing target feature representations from easy-to-hard transfer. 
    Since explicit feature alignment approaches are infeasible for source data-free domain adaptation, we explore the implicit knowledge transfer hidden in the source model from the following two aspects: the curriculum-style feature \textbf{A}lignment and the complementary curriculum-style self-\textbf{T}raining. At the feature alignment stage, we believe that different target samples should be treated unequally during model adaptation because hard samples perform poorly due to domain discrepancy. Explicitly emphasizing these samples with high prediction uncertainty at the early training stage may lead to the convergence issue. To address this issue, we incorporate the idea of Curriculum Learning (CL) \cite{Curriculum,wang2021survey,fan2017self,kumar2010self,ge2020self,peng2021self} into model adaptation. We develop a curriculum-style entropy minimization method to emphasize easy samples first and hard samples later automatically.   
    Through this strategy, we can encourage confident predictions about target predictions. Different from previous curriculum-learning methods \cite{PyCDA,C-SFDA,Curriculum} alleviate domain gaps at the global level, our method focuses on transfer implicit knowledge from easy samples and then gradually propagating this knowledge to more challenging samples. Then, we train the segmentation model to fit target data by the proposed complementary curriculum-style self-training, which includes the negative pseudo-labeling and the positive pseudo-labeling.
    To be specific, the negative pseudo labels refer to the predictions with low confidence scores, providing reliable supervised information to indicate the absent classes for corresponding pixels. Although predictions with negative pseudo labels cannot be identified with correct labels, they can definitely indicate absent classes. 
    Existing LD \cite{LD-SFDA} also utilizes negative labeling to tackle model adaptation, but they treat all negative labels equally during adaptation. In our work, we assign the negative pseudo labels according to the prediction uncertainty, which can reduce the damage of label noise.
    Moreover, to further reduce intra-domain discrepancy within the target domain, we provide an information \textbf{P}ropagation scheme following the implicit knowledge transfer stage.
    We conduct a proxy semi-supervised learning task for target data and propose semantic contrast learning techniques to boost performance. This propagation acts as a generic standard post-processing module for domain adaptation. Furthermore, we also extend a challenging black-box source model scenario, where only the source model’s predictions are available. 
    Our main contributions are summarized as follows: 
    \begin{enumerate}
        \item We propose a novel curriculum-style self-training approach
        for source data-free domain adaptive semantic segmentation, which achieves curriculum-style feature alignment to encourage confident predictions and performs complementary self-training via negative and positive pseudo-labeling. 
        \item We propose an information propagation scheme to further reduce the intra-domain discrepancy within the target domain, which acts as a standard post-processing method for the domain adaptation field. The proposed method also performs well on the black-box source model scenario, where only the source model's predictions are available. 
        \item The proposed approach yields state-of-the-art cross-domain results on synthetic-to-real and normal-to-adverse conditions. Moreover, it performs on par with domain adaptation methods when accessing the source data.
    \end{enumerate}

\section{Related Work}
    \subsection{Unsupervised Domain Adaptation}
    \noindent As a typical field of transfer learning \cite{pan2009survey}, unsupervised domain adaptation (UDA) aims to transfer the knowledge from the labeled source domain to a related but distinct unlabeled target domain. Early UDA methods \cite{zadrozny2004learning,sugiyama2007direct} assume the covariate shift with the identical conditional distributions across domains and reduce the domain discrepancy by estimating the weight of each source instance and re-weighting the source empirical risk. Later, researchers \cite{gong2013connecting,liang2018aggregating} resort to domain-invariant feature learning where a common space with the aligned distributions is learned. However, the transferability of these methods is restricted due to the limited representation ability \cite{long2018transferable}. 
    
    Recently, deep neural networks approaches have achieved inspiring results for unsupervised domain adaptation in many visual applications, for instance, image classification \cite{wang2019transferable,long2018conditional,MSTDA}, face recognition \cite{kan2014domain,guo2021decomposed}, objection detection \cite{yang2021st3d,chen2018domain,vs2021mega}, person re-identification \cite{yang2020part,feng2021complementary,luo2020generalizing,luo2022learning}, and semantic segmentation \cite{AdaptSeg,fan2022toward,IntraDA,ADVENT,UncerDA,ProDA,MMT,wang2022remember,wang2023using}. The structures of these methods usually have three components: deep feature extractor, task-specific classifier, and domain-invariant feature learning module. Based on the strategies of domain-invariant feature learning, existing deep UDA can be roughly divided into three distinct categories: discrepancy-based, adversarial-based, and reconstruction-based. Discrepancy-based approaches aim to minimize the distance between the source and target distributions by utilizing a divergence criterion. 
    As great success has been achieved by GAN method \cite{GAN}, adversarial-based has been a mainstream UDA approach that learns domain-invariant feature representations via confusing source and target domains. DANN \cite{DANN} adopts a gradient reversal layer (GRL) as a domain-invariant feature learning module, ensuring the feature distributions over two domains are made similar. ADDA \cite{ADDA} provides a symmetrical architecture for source and target mapping, which is flexible to learn a powerful feature extractor. Besides, in contrast to the above binary classifier methods, the following works attempt to align joint distribution by considering multiple class-wise domain classifiers \cite{pei2018multi} or a semantic multi-output classifier \cite{cicek2019unsupervised,kurmi2019looking}. Moreover, reconstruction-based approaches \cite{ghifary2016deep} utilize an auxiliary reconstruction task to create a shared domain-invariant representation between two domains and keep the individual characteristics of each domain. In addition, some other reconstruction-based methods \cite{bousmalis2016domain,murez2018image} further improve the adaptation performance by seeking domain-specific reconstruction and cycle consistency. The recent methods \cite{bu2021gaia,peng2023gaia,HPM,DropPos} learn a super pre-trained network to transfer knowledge into the downstream tasks. Beyond them, some other studies also investigate batch normalization \cite{maria2017autodial,wang2019transferable} and adversarial dropout \cite{saito2018adversarial,lee2019drop} techniques to ensure feature alignment.     

    \subsection{Domain Adaptive Semantic Segmentation} 

    \noindent Domain adaptive semantic segmentation (DASS) is a challenging application for UDA, which aims to provide pixel-level predictions for unlabeled target data. Since Hoffman, \textit{et al.} \cite{hoffman2016fcns} introduced DASS, it has achieved much attention due to annotating pixel-level labels being labor-expensive and time-consuming. Existing domain adaptation methods for semantic segmentation can be roughly categorized into two groups: adversarial learning-based methods and self-supervised learning-based methods. For adversarial learning, numerous works focus on reducing the domain discrepancy in the image, feature, and output levels.
    Specifically, image-level methods aim to transfer the domain ``style" (appearance) from target to source \cite{yang2020label}, from source to target \cite{TGCF-DA,UncerDA,yang2021exploring,DCAN,paul2020domain}, or consider both \cite{CyCADA,zhu2018penalizing,CrDoCo,DPL,CAG}. Feature-level based approaches align distributions between the source and target data at different layers of networks by minimizing the feature discrepancy \cite{zhang2020transferring} or adversarial training via a domain classifier \cite{hoffman2016fcns,huang2018domain,huang2020contextual}. Output-level adversarial training is first proposed by \cite{AdaptSeg}, introducing a discriminator to distinguish the predictions obtained from the source or the target domain. Besides, \cite{ADVENT} derives so-called ``weighted self-information maps" for minimizing domain discrepancy, and \cite{IntraDA} diminishes the domain gap by introducing an intra-domain adversarial training process. Moreover, benefiting from the systematically studied recent network architectures, Transformers networks have shown the potential for semantic segmentation, like SegFormer \cite{SegFormer} and P2T \cite{P2T}. Specifically, DAFormer \cite{DAFormer} is the first work to explore a Transformer encoder to the DASS and shows impressive results on the synthetic-to-real benchmarks. MIC \cite{MIC} and related works \cite{DAGFormer,I2F} are the new state-of-the-art approaches by learning spatial context relations of the target domain for robust visual recognition. SePiCo \cite{SePiCo} and T2S-DA \cite{T2S-DA} introduce pixel-level contrastive learning to alleviate slight appearance differences among source and target images.
    
    For the self-supervised learning method, the essential idea is to generate reliable pseudo labels. Typical approaches usually consist of two steps: 1) generate pseudo labels based on the source model \cite{CBST,CRST,IAST} or the learned domain-invariant model \cite{Rectifying,PyCDA,LSE}, 2) refine the target model supervised by the generated pseudo labels \cite{ProDA,UncerDA}. Moreover, \cite{SFD} and \cite{BiMAL} provide a new unaligned score to measure the efficiency of a learned model on a new target domain. Although these methods have achieved promising results, they usually depend heavily on the labeled source data during adaptation.

    \subsection{Source Data-free Domain Adaptation}
    \noindent Source data-free domain adaptation is introduced by Chidlovskii \textit{et al.} \cite{chidlovskii2016domain}, to tackle domain adaptation problems without accessing the original source dataset. To tackle this problem, \cite{li2020model} exploits the pretrained source model to generate target-style samples, and \cite{SHOT,ahmed2021unsupervised} learns a target-specific feature extraction module by implicitly aligning target representations to the source hypothesis. \cite{kundu2020universal} proposes a universal source data-free domain adaptation method when the knowledge of target labels is not available. Moreover, a few recent approaches \cite{AAN,GSFDA,yan2021augmented,C-SFDA,NRC,A2Net,SFDA-DE,liang2023comprehensive} provide model adaptation solutions for classification problems. As for semantic segmentation tasks, model adaptation is also a feasible method to tackle the source data absent setting. Specifically, \cite{SFDA} leverages a generative model to synthesize fake samples to estimate the source distribution and preserve source domain knowledge via knowledge transfer during model adaptation. \cite{UncertaintyReducing} proposes an uncertainty-reducing method to enhance feature representation. \cite{GenAda} and \cite{SFUDA} achieve promising performance via generating a generalized source model trained by data augmentation strategies. To leverage sufficient information for model adaptation, HCL \cite{HCL} explores historical model predictions during adaptation. LD \cite{LD-SFDA} uses positive and negative labels in a class-balanced manner to address ``winner-takes-all" problem. To tackle model adaptation with source data, we offer a simple and effective framework for semantic segmentation tasks, exploiting the target-specific knowledge to intensify the target feature learning. In this paper, we propose a curriculum-learning method for domain adaptive semantic segmentation without source data, which contains curriculum feature alignment, complementary curriculum self-training, and information propagation. 

    Moreover, another setting that has attracted much attention is black-box source-free domain adaptation, which learns target predictions without requiring either source data or source models during adaptation \cite{DINE}. For example, DINE \cite{DINE} focuses on image classification tasks by distilling knowledge via adaptive label smoothing and fine-tuning the distilled model to fit the target distribution. BBM-BTS \cite{BBM-BTS} first utilizing black-box model adaptation on brain tumor segmentation tasks. BiMem\cite{BiMem} focuses on addressing the intrinsic "forgetting" issue in the semantic segmentation task. In this work, we apply the proposed complementary self-training and information propagation into black-box UDA semantic segmentation tasks.

\begin{figure*}[!tbp]
    \centering
    \includegraphics[width=0.93\linewidth, trim=0 0 0 0,clip]{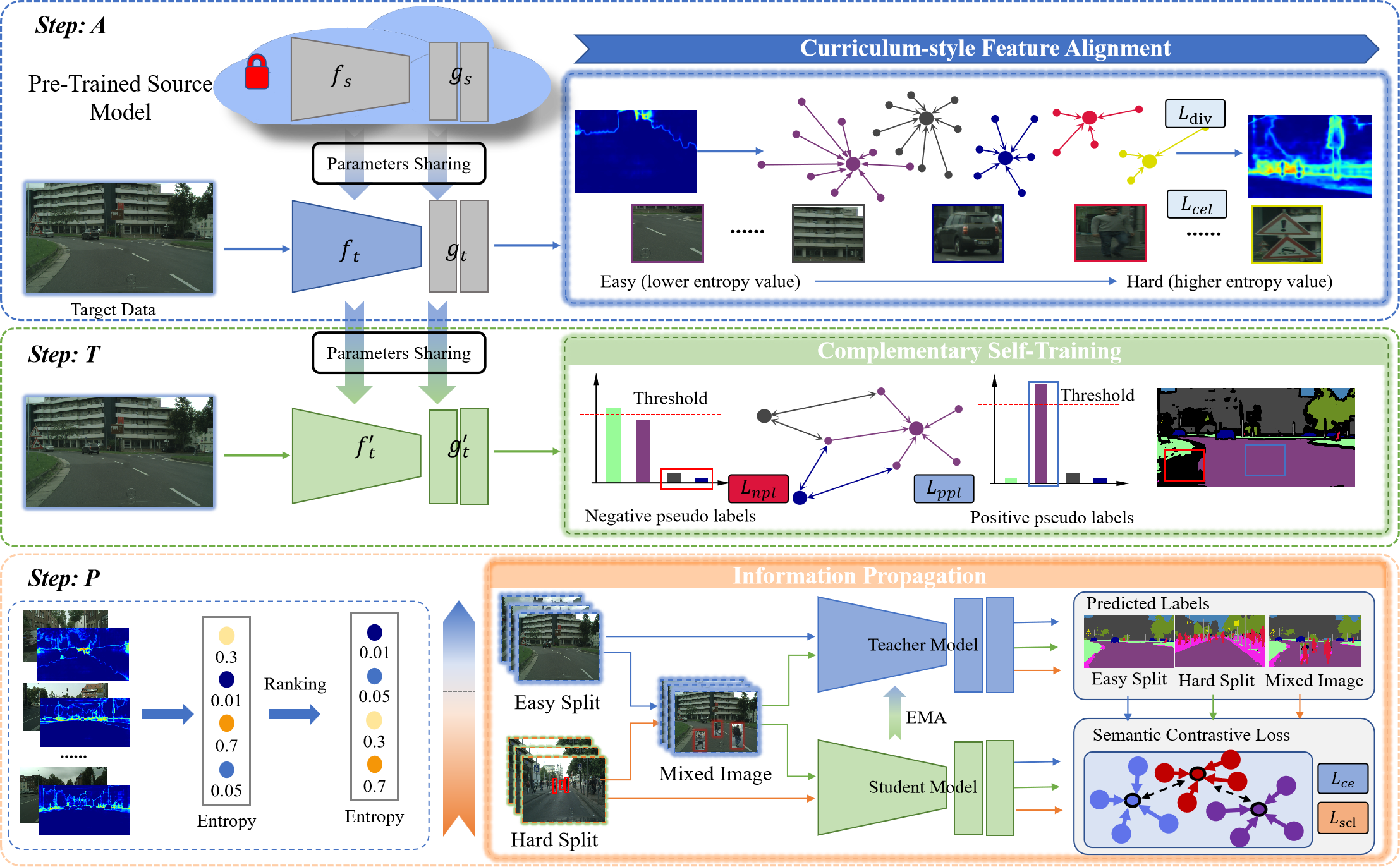} 
    \caption{Illustration of the proposed \textbf{ATP} framework. The step \textbf{\textit{A}} achieves feature alignment via the proposed curriculum-style entropy minimization, which restrains the target prediction from easy-to-hard samples. The step \textbf{\textit{T}} enhances target feature representation learning by the proposed curriculum-style complementary self-training technique, which leverages the negative pseudo labeling and positive pseudo labeling.
    Finally, step \textbf{\textit{P}} achieves an information propagation performing as standard post-processing to reduce the intra-domain discrepancy.} 
    \label{Fig:Pipeline}
    \vspace{-0.3cm}
    \end{figure*}

    \subsection{Semi-supervised Semantic Segmentation} 
    \noindent Semi-supervised learning (SSL) methods exploit a small number of labeled data and a large amount of unlabeled data during training. The key to semi-supervised learning is to learn a consistent representation between the labeled and unlabeled data. To achieve this goal, recent approaches can be divided into two groups: adversarial-based and consistency-based. Inspired by GAN \cite{GAN}, adversarial-based approaches \cite{hung2019adversarial,mittal2019semi,souly2017semi} conduct adversarial loss between labeled and unlabeled data. Consistency-based methods \cite{luo2020semi,ouali2020semi,yuan2021simple} utilize data augmentation and consistency regularization to learn feature representation for unlabeled data. Furthermore, \cite{VAT} first introduces Gaussian perturbations into the input that change output predictions. Then, the strong data augmentation is adopted as solid perturbations. \cite{verma2019interpolation} uses the Mixup as the consistency regularization. \cite{UDA} uses auto-augmentation to generate augmented inputs, and \cite{berthelot2019remixmatch} extends the idea by dividing the augmentation set into strong and weak operations. In this work, we refer to the semi-supervised learning strategy for domain adaptation, aiming to reduce the intra-domain discrepancy within the target domain.

\subsection{Curriculum Learning} 
    \noindent Curriculum learning imitates humans to introduce a learning process from easier samples first and harder samples later, which has been widely used in the domain adaptive semantic segmentation field. Existing methods can be roughly divided into three types from the perspective of ordering tasks, utilizing data, and selecting intermediate domains. The first type \cite{PyCDA,Curriculum} boosts the semantic knowledge transfer by adding easier tasks than semantic segmentation. The second type \cite{CBST,CRST} selects more reliable pseudo labels for target data. The third type \cite{CMA,MA,FoggyDA} conducts the curriculum learning process by gradually inserting intermediate domains, which makes the difficult adaptation problem become multi-step easy adaptation. C-SFDA \cite{C-SFDA} is the most similar curriculum learning work to our method, which focuses on noise-free self-training by exploring the reliability of generated pseudo-labels. Specifically, C-SFDA employs a curriculum learning strategy that prioritizes learning from easy-to-learn samples first and gradually propagates more refined label information among less reliable samples later. CurricuDA \cite{Curriculum} performs curriculum learning from global-to-local label distribution alignment. It first learns the easier global label distributions over images and solves the harder local distributions over landmark superpixels of the target domain. And PyCDA \cite{PyCDA} focuses on properties about the desired label distributions over the target domain images, image regions, and pixels. Different from these works, the proposed ATP integrates the idea of curriculum learning and self-training to tackle source data-free domain adaption. Specifically, it first solves an easy implicit feature alignment task to encourage confident predictions, which starts training the segmentation networks by encouraging reliable samples with confident predictions and then progressively fine-tunes the network to handle unreliable samples with less certain or more challenging predictions. As for the self-training process, the proposed ATP introduces negative pseudo labeling and positive pseudo labeling. 


\section{The Proposed Method}
    \noindent In this section, we introduce the proposed \textbf{ATP} for source data-free domain adaptive semantic segmentation, which aims to explore the implicit knowledge hidden in the source model by the proposed curriculum-style implicit feature alignment and curriculum-style complementary self-training. We also provide information propagation as a standard post-processing for domain adaptation to reduce intra-domain discrepancy. Then, we extend the proposed method to a more challenging source-model black-box case, where only the source model's predictions are available. 

    During training, we are given the source model $\mathcal{M}_s: \mathcal{X}_s \rightarrow \mathcal{Y}_s$ trained on $n_s$ labeled images $\{x_s^i, y_s^i\}_{i=1}^{n_s}$ from the source domain $\mathcal{D}_s$ and $n_t$ unlabeled images $\{x_t^i\}_{i=1}^{n_t}$ from the target domain $\mathcal{D}_{t}$, where $x_s^i \in \mathcal{X}_s$, $y_s^i \in \mathcal{Y}_s$, and $x_t \in \mathcal{X}_t$. Our goal is to learn a segmentation mapping $\mathcal{M}_t: \mathcal{X}_t \rightarrow \mathcal{Y}_t$ transferring from $\mathcal{M}_s$, which predicts a pixel-wise label $y_t^i \in \mathcal{Y}_t$ for a given image $x_t^i$. 

    The semantic segmentation model $\mathcal{M}_s$ on the source domain is obtained in a supervised manner by minimizing the following cross-entropy loss,
    \begin{equation}\label{Eq:ce}
     \mathcal{L}_{ce} = -\frac{1}{n_s}\sum_{i=1}^{n_s}\sum_{j=1}^{H\times W}\sum_{c=1}^{C}y_s^{(i,j,c)}\log p_s^{(i,j,c)},
    \end{equation}
    where $n_s$ is the number of source images, $H$ and $W$ denote the image size, and $C$ denotes the number of categories.
    $p_s^{(i,j,c)}$ denotes the predicted category probability by $\mathcal{M}_s$ and the $y_s^{(i,j,c)}$ is the corresponding one-hot ground-truth label. Generally, the segmentation model $\mathcal{M}_{s}$ consists of a feature extractor $f_s$ and a classifier $g_s$, \textit{i.e.}, $\mathcal{M}_s=f_s \circ g_s$.

    To transfer the trained $\mathcal{M}_{s}$ to the target domain, the proposed \textbf{ATP} framework is shown in Figure \ref{Fig:Pipeline}. We explore the implicit knowledge hidden in the source model from two aspects: First, we freeze the classifier and train the feature extractor to achieve feature alignment via the proposed curriculum-style entropy minimization strategy and weighted diversity loss, which restrains the target prediction from easy-to-hard, referring to Sec. \ref{Sec:3.1}; 
    Second, we enhance target feature representation learning by the proposed complementary self-training technique described in Sec. \ref{Sec:3.2}, which leverages the negative pseudo labeling as well as the class-balanced positive pseudo labeling ($\mathcal{L}_{ppl}$).
   Finally, we achieve an information propagation performing as a standard post-process for domain adaptation to reduce the intra-domain discrepancy by the proposed semantic contrastive loss. 

\subsection{Curriculum Feature Alignment}\label{Sec:3.1}

    \noindent Without source data, explicit feature alignment that directly minimizes the domain gap between the source and target data can not be implemented. Besides, self-training techniques perform poorly because pseudo-labels generated from the source-only model are unreliable. To tackle this problem, we first resort to hypothesis transfer inspired by Liang \textit{et al.} \cite{SHOT} to encourage confident predictions. In contrast, we propose a curriculum-style entropy loss for implicit feature alignment by solving the easy samples first and the hard samples later.    
    
    To eliminate the domain gap, an entropy minimization strategy is adopted for training. Although previous methods \cite{ADVENT,IntraDA} have demonstrated the effectiveness of entropy minimization, they are not appropriate for the source data-free setting because they treat equal importance for different samples. However, hard-to-transfer samples with uncertain predictions (high entropy values) may deteriorate the target feature learning procedure. To address this issue, we attempt to explore more reliable supervision from the easy-to-transfer examples with certain predictions (lower entropy values). Inspired by the curriculum learning algorithm \cite{CBST,Curriculum}, we focus the target data training on automatically emphasizing the easy samples first and hard samples later. As Figure \ref{Fig:Pipeline} shows, easy-to-transfer samples usually have reliable predictions, which benefits the model fitting to the target data at the start. Then, the obtained powerful target model can deal with hard-to-transfer samples, preventing over-fitting occurrence.
    Specifically, we exploit a curriculum-style entropy loss that expects to rapidly focus the model on certain predictions and down-weight the contribution of uncertain predictions. Formally, the curriculum-style entropy loss is formulated as follows:
    \begin{equation}\label{Eq:fel}
       \mathcal{L}_{cel} = \alpha * (1 - h(x_t))^{\gamma} * h(x_t),
    \end{equation}
    where $\alpha$ balances the importance of certainty/uncertainty predictions and $\gamma$ controls the weight of certainty samples. 
    $h(x_t)=-\sum_{c=1}^{C}p_{x_t}^{(h,w,c)}\log p_{x_t}^{(h,w,c)}$ denotes the entropy map and $p_{x_t}^{(h,w,c)}$ is the predicted probability of the target image $x_t$, \textit{i.e.}, $p_{x_t}^{(h,w,c)}=f_t \circ g_t(x_t)$. $f_t$ and $g_t$ denote the feature extractor and classifier for the target data. Weights of hard-to-transfer samples with higher $h(x_t)$ are reduced, and easy samples are emphasized relatively. 
    During training, we pursue hypothesis transfer \cite{SHOT} by fixing the classifier module to implicitly align the target features with the source features, \textit{i.e.}, $g_t = g_s$. As we utilize the same classifier module for different domain-specific features, the optimal target features are enforced to fit the source feature distribution, as they should have a similar one-hot encoding output. Experiments in Sec. \ref{Sec:Exp} reveal this strategy is important.
    
    From the above loss, the trained model concentrates more on easy-to-transfer classes with a sight domain gap or a large pixel proportion, which may lead to a trivial solution with under-fitting hard-to-transfer classes. To tackle this issue, we develop a diversity-promoting loss to ensure the global diversity of the target outputs. Exactly, we expect to see the prediction of the target output containing all categories. The proposed confidence-weighted diversity objective is below,
    \begin{equation}\label{Eq:div}
      \mathcal{L}_{div} = -\sum_{c=1}^{C} \hat{p}_{x_t}^{(h,w,c)}\log \hat{p}_{x_t}^{(h,w,c)},
    \end{equation}
    where $\hat{p}_{x_t}^{(h,w,c)}$ is the weighted mean output embedding of the target image $x_t$. The weight is calculated based on the entropy $h(x_t)$ and $\hat{p}_{x_t}^{(h,w,c)} = \sum_{(h,w)}\exp({-\lambda * h(x_t)})*p_{x_t}^{(h,w,c)}$. $\lambda$ is a hyper-parameter and we empirically set $\lambda=3$ in all experiments. By diversifying the output of target prediction, this technique can circumvent the problem of wrongly predicting confusing target instances as relatively easy classes to learn.

\subsection{Complementary Curriculum  Self-training}\label{Sec:3.2}

\begin{figure}[!tbp]
        \centering
        \includegraphics[width=0.99\linewidth]{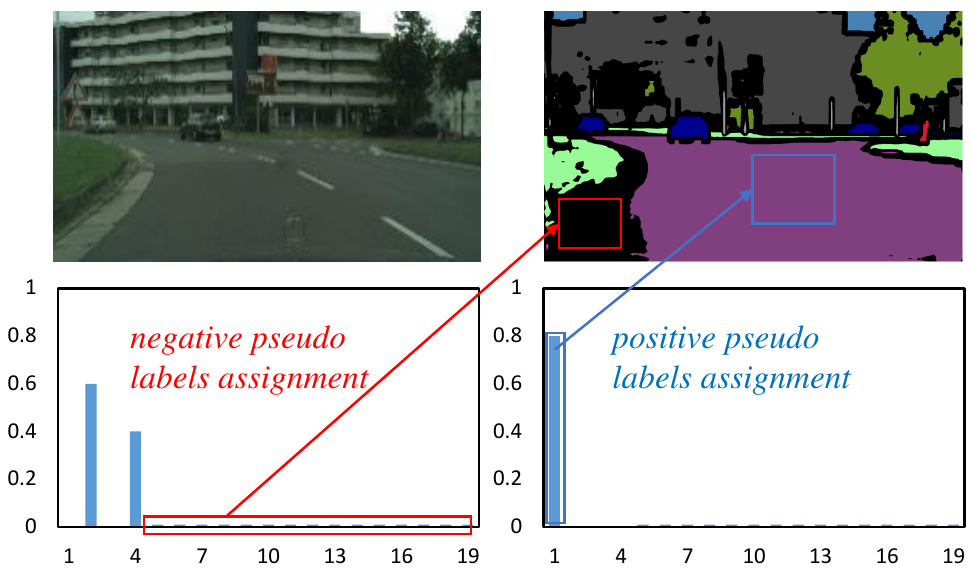}
        \caption{Complementary pseudo labels. Blue refers to positive pseudo labels, and red refers to negative pseudo labels.}
        \label{Fig:NPL}
        \vspace{-0.4cm}
    \end{figure}

    \noindent\textbf{Negative pseudo labeling.} 
    Previous self-supervised learning methods focus on strengthening the reliability of pseudo labels by developing various denoising strategies \cite{ProDA, UncerDA}, but they ignore most of the pixels with lower prediction confidence scores. In that case, the model tends to be over-fitting because the selected over-confident samples lack references using negative pseudo labels. To remedy this problem, we restrict our attention to the predictions with lower confidence values, termed negative pseudo labels. Although low-confidence predictions cannot be assigned as the correct labels, they will definitely indicate specific absent classes. As Figure \ref{Fig:NPL} shows, it is hard to clearly indicate which category it belongs to for the ignore pixels (black areas), but it is easy to determine which categories it does not belong to.

    For the negative pseudo-label assignment, our goal is to explore the implicit knowledge in uncertainty-aware regions that have been ignored in previous works. Although these predictions have higher uncertainty due to the uniform or low prediction scores, they can definitely point out the absent categories. For example, for a pixel whose output is [0.48, 0.47, 0.02, 0.03], it is impossible to determine which category it belongs to, but we can confidently point out that it does not belong to classes with scores of 0.02 and 0.03. Therefore, we use all absent classes as supervision information for target feature learning. Formally, we assign the negative pseudo labels as follows:
    \begin{equation}\label{Eq:NPL}
      \delta(p_{x_t}^{(h,w,c)}) =
      \begin{cases}
      1  & \mathrm{if} \  p_{x_t}^{(h,w,c)} < \lambda_{neg}, \\
      0  & \mathrm{otherwise},
       \end{cases}
    \end{equation}
    where $\lambda_{neg}$ is the negative threshold and we use $\lambda_{neg} = 0.05$ for all experiments. It is worth noticing that the negative pseudo labels are binary multi-labels, which certainly indicates the absence of relative classes. The negative pseudo-labeling objective is formulated as follows: 
    \begin{equation}\label{Eq:NPLL}
       \mathcal{L}_{npl} = -\frac{1}{n_t}\sum_{i=1}^{n_t}\sum_{j=1}^{H\times W}\sum_{c=1}^{C}\delta(p_{x_t}^{(i,j,c)}) \log (1-p_{x_t}^{(i,j,c)}).
    \end{equation}
    Intuitively, negative pseudo labeling is an easier task than positive pseudo labeling because it does not need to identify the correct category. We then solve the hard positive pseudo labeling by tackling samples from easy to hard.

    \textbf{Positive pseudo labeling.} We then introduce the positive pseudo labeling method, which aims to select reliable pseudo labels with high confidence scores first and then propagate the learned knowledge to hard predictions. Considering the domain-transfer difficulty among different categories, generating pseudo labels according to high confidence will cause hard-to-transfer categories to be ignored. Because easy-to-transfer classes (\textit{e.g.}, roads, buildings, walls, sky) usually have higher prediction confidence scores, while hard-to-transfer classes (\textit{e.g.}, lights, signals, trains, bicycles) perform poorly. 
    To balance these easy and hard samples, we assign pseudo-labels based on the guidance of the category-level threshold inspired by \cite{CBST} as below,
    \begin{equation}\label{Eq:PPL}
    \hat{y}_{x_t}^{(h,w,c)} =
    \begin{cases}
    1 & \mathrm{if} \ c = (\mathrm{argmax}_{c}p_{x_t}^{(h,w,c)})  \cap
     (p_{x_t}^{(h,w,c)} > \lambda_{c})  \\
    0 & \mathrm{otherwise} ,
    \end{cases}
    \end{equation}
    where $\lambda_c$ is a category-level threshold and it is dynamically increased for propagating to the unreliable predictions.

    Then, we optimize the target model by minimizing the categorical cross-entropy with every pseudo label $\hat{y}_{t}$:
    \begin{equation}\label{Eq:PPLL}
      \mathcal{L}_{ppl} = -\frac{1}{n_t}\sum_{i=1}^{n_t}\sum_{j=1}^{H\times W}\sum_{c=1}^{C}\hat{y}_t^{(i,j,c)}\log p_{x_t}^{(i,j,c)},
    \end{equation}
    where $n_t$ is the number of target images, $p_{x_t}^{(i,j,c)}$ denotes the predicted category probability, and $\hat{y}_t^{(i,j,c)}$ is the corresponding pseudo label obtained from Eq.~(\ref{Eq:PPL}).

    Finally, the complementary self-training objective includes positive and negative pseudo labeling:
    \begin{equation}\label{Eq:PNL}
       \mathcal{L}_{cst} = \mathcal{L}_{ppl} + \mathcal{L}_{npl}.
    \end{equation}

\subsection{Information Propagation}\label{Sec:3.3}

    \noindent Although we propose a much better solution for global adaptation without source data, intra-domain discrepancy still exists, as previous works \cite{IntraDA,SFDA} indicate. To address this issue, we first divide the target data into easy and hard splits and then close the intra-domain gap to boost the adaptation performance. Contrary to the traditional adversarial mechanism, we borrow a semi-supervised learning scheme by considering the easy split as labeled data and the hard split as unlabeled. The target domain is separated using the entropy ranking strategy as below,
    \begin{equation}\label{Eq:rank}
        r(x_t) = \frac{1}{HW} \sum_{h=1}^{H}\sum_{w=1}^{W}\mathcal{H}(x_t)^{(h,w)},
    \end{equation}
    which is the mean entropy for a target image $x_t$ with the corresponding entropy map $\mathcal{H}(x_t)$. Easy and hard splits are conducted from the ranking of $r(x_t)$ following $ratio = \frac{\|x_{te}\|}{(\|x_{te}\| + \|x_{th}\|)}$, where $x_{te}$ and $x_{th}$ denote the easy split and the hard split. Details are shown in Figure \ref{Fig:Pipeline}. 

    After that, we conduct standard post-processing for domain adaptation following a semi-supervised learning manner based on the assumption that pseudo labels of the easy part are reliable. Specifically, we incorporate an augmentation strategy in the hard split and conduct a semantic contrastive loss between easy and hard split to regularize the semi-supervised learning. We employ ClassMix \cite{ClassMix} technique as the augmentation strategy, and the semantic contrastive loss is formalized as:
    \begin{equation}\label{Eq:SCL}
        \begin{split}
            \mathcal{L}_{scl} = - \sum_{c \in C} \sum_{r_{th}^{c} \sim \mathcal{R}_{th}^{c}} \log \frac{\exp(r_{th}^{c} * r_{te}^{c}/\tau)}{\sum_{r_{te}^{c}\sim \mathcal{R}_{te}^{c}}\exp(r_{th}^{c} * r_{te}^{c} /\tau)} 
        \end{split}
    \end{equation}
    where $r_{th}^c$ denotes the normalized presentation of the hard split image with class $c$ and $r_{te}^{c}$ is the easy split target image whose label does not belong to class $c$. $\mathcal{R}_{th}^{c}$ and $\mathcal{R}_{te}^{c}$ represent the corresponding set of target image. $\tau$ is the temperature. The pseudo labels are generated from the teacher model, and parameters are updated following the exponential moving average (EMA) strategy according to the target student model $\mathcal{M}_t$ without gradient optimization.
    
    Then, the final proposed information propagation objective is as follows:
    \begin{equation}\label{Eq:divideMix}
    \begin{split}
     \mathcal{L}_{ssl} =  & \mathcal{L}_{ce}(\mathcal{M}_{t}(x_{te}),\hat{y}_{te})  + \mathcal{L}_{scl}        
    \end{split}
    \end{equation}
    where $\mathcal{L}_{ce}$ is the cross-entropy loss as Eq.~(\ref{Eq:ce}) describes. $\hat{y}_{te}$ denotes the pseudo labels of $x_{te}$ obtained from the trained model in Eq.~(\ref{Eq:PNL}). The detailed process is shown in Figure \ref{Fig:Pipeline}.

    \begin{algorithm}[!t]
    \caption{\textbf{Curriculum-style Self-training}}
    \label{algorithm1}
    \SetAlgoLined
    \KwData{training dataset: $\mathcal{X}_t$; the trained source model: $\mathcal{M}_{s}$; parameters: $\alpha$, $\gamma$, $\lambda_{c}$, and $\lambda_{neg}$;}
    \KwResult{the output target model $\mathcal{M}_{t_1}$}
   
    Warmup: $\mathcal{M}_{s}\gets(\mathcal{X}_{s},\mathcal{Y}_{s})$ according to Eq.(\ref{Eq:ce}), which is trained before;
    
    \
    \textbf{Curriculum Feature Alignment (A):} 
  
    initialization: 
    $\mathcal{M}_{t_0}=f_{t_0} \circ g_{t_0} \gets \mathcal{M}_s$. $f_{t_{0}}$ is optimized and $g_{t_0}$ is frozen.
  
    \For{$m \gets 0$ \KwTo $epochs$}{
        Optimize $\mathcal{M}_{t_0}$ by minimizing the proposed losses $\mathcal{L}_{cel}$ Eq. (\ref{Eq:fel}) and $\mathcal{L}_{div}$ Eq. (\ref{Eq:div});
    }
    \ 
    \textbf{Curriculum Complementary Self-training (T):}
  
    initialization: $\mathcal{M}_{t_1} \gets \mathcal{M}_{t_0}$;\\
    $\mathcal{P}$ is a set of positive prediction samples; $\mathcal{N}$ is a set of negative prediction samples;\\
    \For{$stage \gets 1$ \KwTo $S$} {
        Generate positive pseudo labels set $\mathcal{P} \gets \mathcal{M}_{t_{1}}(\mathcal{X}_t; \lambda_c)$ according to Eq. (\ref{Eq:PPL}); \\
    \For{$m \gets 0$ \KwTo $epochs$}{
        Get negative pseudo labels $\mathcal{N} \gets \mathcal{M}_{t_1}(\mathcal{X}_{t};\lambda_{neg})$ according to Eq. (\ref{Eq:NPL}); \\
        Train model $\mathcal{M}_{t_1}$ using the proposed complementary self-training technique according to Eq. (\ref{Eq:PNL}); \\
      }
  } 
  
    \textbf{Return} $\mathcal{M}_{t_1}$
    \end{algorithm}

    Unlike the most closely related work \cite{IntraDA} that reduces the intra-domain gap by conducting adversarial learning, our work directly learns semantic consistency representations for the target data in a semi-supervised learning manner. Notice that the adopted technique can be considered as standard post-processing for adaptation, which is more stable than the adversarial learning method.

    \begin{algorithm}[!t]
    \caption{\textbf{Information Propagation (P)}}
    \label{algorithm}
    \SetAlgoLined
    \KwData{training dataset: $\mathcal{X}_t$; the trained target model: $\mathcal{M}_{t_1}$; parameters: $ratio$}
    \KwResult{the output target model $\mathcal{M}_{t}$}
   
    Initialization: $\mathcal{M}_t \gets \mathcal{M}_{t_1}$; \\
    Divide the target data $\mathcal{X}_t$ into easy split $\mathcal{X}_{te}$ and hard split $\mathcal{X}_{th}$ according to Eq. (\ref{Eq:rank}); \\ 
    Generate pseudo labels $\mathcal{\hat{Y}}_{te}$ for easy split $\mathcal{X}_{te}$; \\
    \For{$m \gets 0$ \KwTo $epochs$}{
  
    Apply data augmentation for the hard target data $x_{th}$; \\
    Optimize $\mathcal{M}_{t}$ by minimizing the proposed loss $\mathcal{L}_{ssl}$ according to Eq. (\ref{Eq:divideMix});
  }
  
    \textbf{Return} $\mathcal{M}_{t}$
   
   \end{algorithm}                                                                                             
    \begin{figure}[!t]
        \centering
        \includegraphics[width=0.99\linewidth,trim=170 120 180 130,clip]{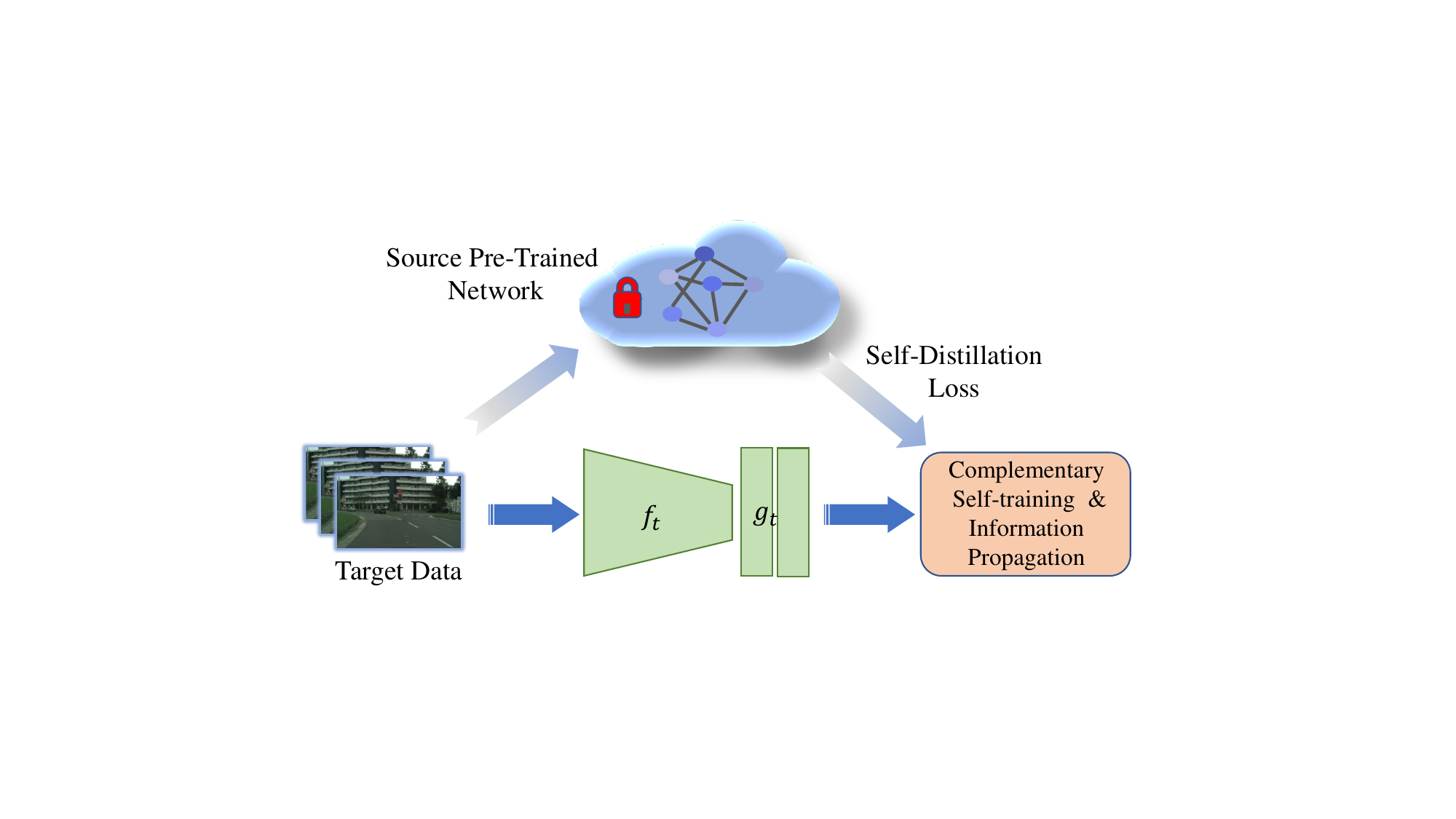}
        \caption{Illustration of the black-box source-model scenario.}
        \label{Fig:BlackBox}
        \vspace{-0.4cm}
    \end{figure}


    \begin{table*}[!t]
     \centering
     \tabcolsep=4pt
     \caption{Semantic segmentation performance of GTA5 $\rightarrow$ Cityscapes (CS) trained on different ResNet-101, P2T-Base, and MiT-B5 backbones. ``SF" denotes the source data-free setting. 
     }\label{Tab:GTA2CS}
     \resizebox{\textwidth}{!}{
     \begin{tabular}{lccccccccccccccccccccc}
       \toprule[1.0pt]
        {Method} & {SF} & \rotatebox{90}{road} & \rotatebox{90}{side.} & \rotatebox{90}{build.} & \rotatebox{90}{wall}   & \rotatebox{90}{fence} & \rotatebox{90}{pole} & \rotatebox{90}{light}  & \rotatebox{90}{sign}  & \rotatebox{90}{vege.} & \rotatebox{90}{terr.}  & \rotatebox{90}{sky}   & \rotatebox{90}{person} & \rotatebox{90}{rider}  & \rotatebox{90}{car}   & \rotatebox{90}{truck} & \rotatebox{90}{bus}    & \rotatebox{90}{train} & \rotatebox{90}{motor.} & \rotatebox{90}{bike}   & \rotatebox{90}{\textbf{mIoU}} \\

      \midrule[1.0pt]
      \multicolumn{22}{c}{\textbf{ResNet-101}} \\
      \midrule[0.7pt]


       AdaptSeg \cite{AdaptSeg} & \XSolidBrush & 86.5 & 36.0 & 79.9 & 23.4 & 23.3 & 23.9 & 35.2 & 14.8 & 83.4 & 33.3 & 75.6 & 58.5 & 27.6 & 73.7 & 32.5 & 35.4 & 3.9 & 30.1 & 28.1 & 42.4 \\
       ADVENT \cite{ADVENT} & \XSolidBrush & 89.4 & 33.1 & 81.0 & 26.6 & 26.8 & 27.2 & 33.5 & 24.7 & 83.9 & 36.7 & 78.8 & 58.7 & 30.5 & 84.8 & 38.5 & 44.5 & 1.7 & {31.6} & 32.4 & 45.5 \\
       CBST \cite{CBST} & \XSolidBrush & 91.8 & {53.5} & 80.5 & 32.7 & 21.0 & 34.0 & 28.9 & 20.4 & 83.9 & 34.2 & 80.9 & 53.1 & 24.0 & 82.7 & 30.3 & 35.9 & 16.0 & 25.9 & 42.8 & 45.9 \\
       MaxSquare \cite{MaxSquare} & \XSolidBrush & 89.4 & 43.0 & 82.1 & 30.5 & 21.3 & 30.3 & 34.7 & 24.0 & 85.3 & 39.4 & 78.2 & 63.0 & 22.9 & 84.6 & 36.4 & 43.0 & 5.5 & 34.7 & 33.5 & 46.4 \\
       CAG\_UDA \cite{CAG} & \XSolidBrush & 90.4 & 51.6 & 83.8 & 34.2 & 27.8 & {38.4} & 25.3 & {48.4} & 85.4 & 38.2 & 78.1 & 58.6 & {34.6} & 84.7 & 21.9 & 42.7 & {41.1} & 29.3 & 37.2 & 50.2 \\
       FDA \cite{FDA} & \XSolidBrush & {92.5} & 53.3 & 82.4 & 26.5 & 27.6 & 36.4 & {40.6} & 38.9 & 82.3 & 39.8 & 78.0 & 62.6 & 34.4 & 84.9 & 34.1 & {53.1} & 16.9 & 27.7 & {46.4} & {50.5} \\
       PLCA \cite{PLCA} & \XSolidBrush & 84.0 & 30.4 & 82.4 & {35.3} & 24.8 & 32.2 & 36.8 & 24.5 & {85.5} & 37.2 & 78.6 & {66.9} & 32.8 & {85.5} & 40.4 & 48.0 & 8.8 & 29.8 & 41.8 & 47.7 \\
       SIM \cite{SIM} & \XSolidBrush & 90.6 & 44.7 & {84.8} & 34.3 & {28.7} & 31.6 & 35.0 & 37.6 & 84.7 & {43.3} & {85.3} & 57.0 & 31.5 & 83.8 & {42.6} & 48.5 & 1.9 & 30.4 & 39.0 & 49.2 \\

       \midrule[0.7pt]
       SFDA \cite{SFDA} & \Checkmark & 84.2 & 39.2 & 82.7 & 27.5 & 22.1 & 25.9 & 31.1 & 21.9 & 82.4 & 30.5 & 85.3 & 58.7 & 22.1 & 80.0 & 33.1 & 31.5 & 3.6 & 27.8 & 30.6 & 43.2 \\
       URMA \cite{UncertaintyReducing} & \Checkmark & 92.3 & 55.2 & 81.6 & 30.8 & 18.8 & 37.1 & 17.7 & 12.1 & 84.2 & 35.9 & 83.8 & 57.7 & 24.1 & 81.7 & 27.5 & 44.3 & 6.9 & 24.1 & 40.4 & 45.1 \\
       S4T \cite{S4T} & \Checkmark & 89.7 & 39.2 & 84.4 & 25.7 & {29.0} & {39.5} & {45.1} & 36.8 & {86.8} & 41.8 & 79.3 & 61.2 & 26.7 & 85.0 & 19.3 & 28.2 & 5.3 & 11.8 & 9.3 & 44.8 \\ 
       SFUDA \cite{SFUDA} & \Checkmark & 95.2 & 40.6 & 85.2 & 30.6 & 26.1 & 35.8 & 34.7 & 32.8 & 85.3 & 41.7 & 79.5 & 61.0 & {28.2} & {86.5} & {41.2} & 45.3 & {15.6} & 33.1 & 40.0 & 49.4 \\
       HCL \cite{HCL} & \Checkmark & 92.0 & 55.0 & 80.4 & 33.5 & 24.6 & 37.1 & 35.1 & 28.8 & 83.0 & 37.6 & 82.3 & 59.4 & 27.6 & 83.6 & 32.3 & 36.6 & 14.1 & 28.7 & 43.0 & 48.1 \\ 
       
       LD \cite{LD-SFDA} & \Checkmark & 91.6 & 53.2 & 80.6 & 36.6 & 14.2 & 26.4 & 31.6 & 22.7 & 83.1 & 42.1 & 79.3 & 57.3 & 26.6 & 82.1 & 41.0 & 50.1 & 0.3 & 25.9 & 19.5 & 45.5 \\
       DTAC \cite{DTAC} & \Checkmark & 78.0 & 29.5 & 83.0 & 29.3 & 21.0 & 31.8 & 38.1 & 33.1 & 83.8 & 39.2 & 80.8 & 61.0 & 30.0 & 83.9 & 26.1 & 40.4 & 1.9 & 34.2 & 43.7 & 45.7 \\
       Cal-SFDA \cite{Cal-SFDA} & \Checkmark & 90.0 & 48.4 & 83.2 & 35.5 & 23.6 & 30.8 & 39.6 & 35.9 & 84.3 & 43.2 & 85.1 & 60.2 & 27.9 & 84.3 & 32.6 & 44.7 & 2.2 & 19.9 & 42.1 & 48.1 \\
       C-SFDA \cite{C-SFDA} & \Checkmark & 90.4 & 42.2 & 83.2 & 34.0 & 29.3 & 34.5 & 36.1 & 38.4 & 84.0 & 43.0 & 75.6 & 60.2 & 28.4 & 85.2 & 33.1 & 46.4 & 3.5 & 28.2 & 44.8 & 48.3 \\
       DT-ST \cite{DT-ST} & \Checkmark & 90.3 & 47.8 & 84.3 & 38.8 & 22.7 & 32.4 & 41.8 & 41.2 & 85.8 & 42.5 & 87.8 & 62.6 & 37.0 & 82.5 & 25.8 & 32.0 & 29.8 & 48.0 & 56.9 & 52.1 \\ 
       IAPC \cite{IAPC} & \Checkmark & 90.9 & 36.5 & 84.4 & 36.1 & 31.3 & 32.9 & 39.9 & 38.7 & 84.3 & 38.6 & 87.5 & 58.6 & 28.8 & 84.3 & 33.8 & 49.5 & 0.0 & 34.1 & 47.6 & 49.4 \\
       AUGCO \cite{AUGCO} & \Checkmark & 92.6 & {84.6} & {86.8} & 84.8 & 56.1 & 82.2 & 45.3 & 63.4 & 43.8 & 28.7 & 26.8 & 14.0 & 41.1 & 34.9 & 16.7 & 29.6 & 45.7 & 5.6 & 12.6 & 47.1 \\
       
       \midrule[0.7pt]

       \rowcolor{gray!40} \bf{ATP (w/o TP)} & \Checkmark & 90.1 & 33.7 & 83.7 & 31.5 & 20.6 & 31.7 & 34.5 & 25.4 & 84.8 & 38.2 & 82.0 & 59.9 & 26.9 & 84.5 & 25.9 & 38.9 & 6.9 & 23.8 & 32.1 & 45.0 \\ 
       \rowcolor{gray!40} \bf{ATP} (w/o P) & \Checkmark & 90.4 & 43.9 & 85.1 & 40.4 & 23.8 & 34.2 & 41.8 & 34.3 & 84.2 & 35.9 & 85.5 & 62.3 & 21.9 & 83.8 & 36.9 & 51.8 & 0.0 & 36.8 & 53.0 & 49.8 \\
       \rowcolor{gray!40} \bf{ATP} & \Checkmark & 93.2 & 55.8 & {86.5} & {45.2} & 27.3 & 36.6 & 42.8 & 37.9 & 86.0 & 43.1 & {87.9} & {63.5} & 15.3 & 85.5 & {41.2} & {55.7} & 0.0 & {38.1} & {57.4} & {52.6} \\

       \midrule[0.7pt]
       \multicolumn{22}{c}{\textbf{P2T-Base}} \\
       \midrule[0.7pt]

       P2T \cite{P2T} & \Checkmark & 87.1 & 25.2 & 78.3 & 28.0 & 23.0 & 27.9 & 31.5 & 18.5 & 83.5 & 39.2 & 81.8 & 56.2 & 20.2 & 87.0 & 42.7 & 43.8 & 16.1 & 27.5 & 22.5 & 44.2 \\
       \rowcolor{gray!40} \textbf{ATP (w/o TP)} & \Checkmark & 90.2 & 32.2 & 83.1 & 32.4 & 25.6 & 34.3 & 38.6 & 25.7 & 86.5 & 43.6 & 77.2 & 61.2 & 24.2 & 89.2 & 53.8 & 52.2 & 5.2 & 29.6 & 32.7 & 48.3 \\
       \rowcolor{gray!40} \textbf{ATP (w/o P)} & \Checkmark & 92.9 & 53.2 & 86.6 & 35.7 & 32.9 & 44.3 & 45.0 & 36.9 & 88.1 & 49.4 & 81.0 & 63.7 & 26.5 & 89.5 & 52.6 & 54.7 & 7.2 & 28.1 & 35.1 & 52.8 \\
       \rowcolor{gray!40} \textbf{ATP} & \Checkmark & 94.0 & 59.1 & 87.6 & 35.3 & 40.3 & 42.2 & 45.1 & 47.8 & 88.4 & 53.1 & 84.9 & 67.9 & 29.8 & 89.1 & 56.8 & 56.0 & 10.5 & 31.1 & 41.2 & 55.8 \\

      \midrule[0.7pt]
      \multicolumn{22}{c}{\textbf{MiT-B5}} \\
      \midrule[0.7pt]
      Source Only & - & 77.1 & 15.2 & 83.8 & 30.8 & 32.0 & 27.9 & 41.5 & 18.5 & 86.5 & 42.5 & 86.8 & 62.6 & 22.2 & 87.0 & 42.7 & 36.8 & 6.1 & 33.5 & 12.5 & 44.5 \\
      TransDA-B \cite{TransDA} & \XSolidBrush & 94.7 & 64.2 & 89.2 & 48.1 & 45.8 & 50.1 & 60.2 & 40.8 & 90.4 & 50.2 & 93.7 & 76.7 & 47.6 & 92.5 & 56.8 & 60.1 & 47.6 & 49.6 & 55.4 & 63.9 \\
      DAFormer \cite{DAFormer} & \XSolidBrush & 95.7 & 70.2 & 89.4 & 53.5 & 48.1 & 49.6 & 55.8 & 59.4 & 89.9 & 47.9 & 92.5 & 72.2 & 44.7 & 92.3 & 74.5 & 78.2 & 65.1 & 55.9 & 61.8 & 68.3 \\
      HRDA \cite{HRDA} & \XSolidBrush & 96.4 & 74.4 & 91.0 & 61.6 & 51.5 & 57.1 & 63.9 & 69.3 & 91.3 & 48.4 & 94.2 & 79.0 & 52.9 & 93.9 & 84.1 & 85.7 & 75.9 & 63.9 & 67.5 & 73.8 \\
      IDM \cite{IDM} & \XSolidBrush & 97.2 & 77.1 & 89.8 & 51.7 & 51.7 & 54.5 & 59.7 & 64.7 & 89.2 & 45.3 & 90.5 & 74.2 & 46.6 & 92.3 & 76.9 & 59.6 & 81.2 & 57.3 & 62.4 & 69.5 \\

      \midrule
      DAFormer \cite{DAFormer} & \Checkmark & 87.7 & 33.4 & 83.9 & 28.1 & 27.5 & 35.9 & 42.9 & 28.7 & 82.4 & 28.6 & 83.1 & 65.0 & 37.0 & 85.8 & 53.9 & 46.3 & 31.8 & 23.6 & 36.8 & 49.6 \\
      HRDA \cite{HRDA} & \Checkmark & 83.3 & 28.2 & 83.3 & 43.3 & 22.2 & 42.9 & 47.7 & 38.2 & 87.2 & 40.0 & 81.6 & 69.5 & 35.9 & 84.8 & 42.7 & 50.4 & 41.2 & 33.7 & 29.6 & 51.9 \\
      IDM \cite{IDM} & \Checkmark & 93.9 & 59.1 & 86.6 & 35.3 & 30.4 & 42.2 & 45.1 & 57.8 & 88.4 & 35.1 & 89.4 & 69.7 & 39.8 & 89.1 & 66.8 & 46.0 & 13.5 & 41.1 & 61.2 & 57.4 \\
      ELR \cite{ELR} & \Checkmark & - & - & - & - & - & - & - & - & - & - & - & - & - & - & - & - & - & - & - & 54.1 \\
      UVPT \cite{UVPT} & \Checkmark & - & - & - & - & - & - & - & - & - & - & - & - & - & - & - & - & - & - & - & 56.1 \\

       \midrule

       \rowcolor{gray!40} \textbf{ATP (w/o TP)} & \Checkmark & 93.0 & 54.5 & 87.6 & 32.9 & 27.6 & 42.5 & 43.8 & 55.4 & 87.2 & 31.5 & 89.4 & 65.0 & 34.1 & 85.9 & 50.6 & 45.9 & 44.9 & 16.5 & 37.7 & 54.0 \\
       \rowcolor{gray!40} \textbf{ATP (w/o P)} & \Checkmark & 96.7 & 75.0 & 88.9 & 36.4 & 34.9 & 43.2 & 48.9 & 60.8 & 89.5 & 47.4 & 93.8 & 69.7 & 38.6 & 89.5 & 67.0 & 59.4 & 29.1 & 51.1 & 65.5 & 62.4 \\
       \rowcolor{gray!40} \textbf{ATP} & \Checkmark & 96.6 & 75.3 & 89.4 & 50.2 & 41.5 & 47.5 & 48.6 & 61.1 & 89.8 & 48.3 & 93.4 & 70.4 & 40.1 & 89.8 & 66.7 & 58.2 & 30.3 & 53.4 & 65.6 & 64.0 \\

      \bottomrule[1.0pt]
      \end{tabular}
      }
      \vspace{-0.3cm}
    \end{table*}


    \begin{table*}[!tbp]
     \centering
     \tabcolsep=5.8pt
     \footnotesize
     \caption{Semantic segmentation performance of SYNTHIA $\rightarrow$ Cityscapes (CS) trained on different ResNet-101, P2T-Base, and MiT-B5 backbones. ``SF" denotes the source data-free setting. 
     }\label{Tab:SYN2CS}
     \resizebox{\textwidth}{!}{
     \begin{tabular}{lccccccccccccccccccc}
     \toprule[1.0pt]
     Method & SF & \rotatebox{90}{road} & \rotatebox{90}{side.} & \rotatebox{90}{build.}  &  \rotatebox{90}{wall$^{*}$}  &  \rotatebox{90}{fence$^{*}$}  &  \rotatebox{90}{pole$^{*}$} &  \rotatebox{90}{light}  & \rotatebox{90}{sign}  & \rotatebox{90}{vege.} & \rotatebox{90}{sky}   & \rotatebox{90}{person} & \rotatebox{90}{rider}  & \rotatebox{90}{car}   & \rotatebox{90}{bus}   & \rotatebox{90}{motor.} & \rotatebox{90}{bike}   & \rotatebox{90}{\textbf{mIoU}$^{*}$} & \rotatebox{90}{\textbf{mIoU}} \\

     \midrule[0.7pt]
     \multicolumn{20}{c}{\textbf{ResNet-101}} \\
     \midrule[0.7pt]   


     AdaptSeg \cite{AdaptSeg} & \XSolidBrush & 79.2 & 37.2 & 78.8 & - & - & - & 9.9 & 10.5 & 78.2 & 80.5 & 53.5 & 19.6 & 67.0 & 29.5 & 21.6 & 31.3 & - & 45.9 \\
     ADVENT \cite{ADVENT} & \XSolidBrush & 85.6 & 42.2 & 79.7 & 8.7 & 0.4 & 25.9 & 5.4 & 8.1 & 80.4 & 84.1 & 57.9 & 23.8 & 73.3 & 36.4 & 14.2 & 33.0 & 41.2 & 48.0 \\
     CBST \cite{CBST} & \XSolidBrush & 68.0 & 29.9 & 76.3 & 10.8 & {1.4} & {33.9} & 22.8 & {29.5} & 77.6 & 78.3 & 60.6 & 28.3 & 81.6 & 23.5 & 18.8 & 39.8 & 42.6 & 48.9 \\
     MaxSquare \cite{MaxSquare} & \XSolidBrush & 82.9 & 40.7 & 80.3 & 10.2 & 0.8 & 25.8 & 12.8 & 18.2 & 82.5 & 82.2 & 53.1 & 18.0 & 79.0 & 31.4 & 10.4 & 35.6 & 41.4 & 48.2 \\
     CAG\_UDA \cite{CAG} & \XSolidBrush & {84.8} & 41.7 & {85.8} & - & - & - & 13.7 & 23.0 & {86.5} & 78.1 & {66.3} & 28.1 & 81.8 & 21.8 & 22.9 & 49.0 & - & 52.6 \\
     PLCA \cite{PLCA} & \XSolidBrush & 82.6 & 29.0 & 81.0 & {11.2} & 0.2 & 33.6 & {24.9} & 18.3 & 82.8 & 82.3 & 62.1 & 26.5 & {85.6} & {48.9} & 26.8 & {52.2} & {46.8} & {54.0} \\
     SIM \cite{SIM} & \XSolidBrush & 83.0 & {44.0} & 80.3 & - & - & - & 17.1 & 15.8 & 80.5 & 81.8 & 59.9 & {33.1} & 70.2 & 37.3 & 28.5 & 45.8 & - & 52.1 \\
     FDA \cite{FDA} & \XSolidBrush &  79.3 & 35.0 & 73.2 & - & - & - & 19.9 & 24.0 & 61.7 & {82.6} & 61.4 & 31.1 & 83.9 & 40.8 & {38.4} & 51.1 & - & 52.5 \\

     \midrule[0.7pt]
     SFDA \cite{SFDA} & \Checkmark & 81.9 & 44.9 & 81.7 & 4.0 & 0.5 & 26.2 & 3.3 & 10.7 & 86.3 & {89.4} & 37.9 & 13.4 & 80.6 & 25.6 & 9.6 & 31.3 & 39.2 & 45.9 \\
     URMA \cite{UncertaintyReducing} & \Checkmark &  59.3 & 24.6 & 77.0 & {14.0} & {1.8} & 31.5 & 18.3 & {32.0} & 83.1 & 80.4 & 46.3 & 17.8 & 76.7 & 17.0 & 18.5 & 34.6 & 39.6 & 45.0 \\
     S4T \cite{S4T} & \Checkmark & 84.9 & 43.2 & 79.5 & 7.2 & 0.3 & 26.3 & 7.8 & 11.7 & 80.7 & 82.4 & 52.4 & 18.7 & 77.4 & 9.6 & 9.5 & 37.9 & 39.3 & 45.8 \\
     SFUDA \cite{SFUDA} & \Checkmark & {90.9} & 45.5 & 80.8 & 3.6 & 0.5 & 28.6 & 8.5 & 26.1 & 83.4 & 83.6 & 55.2 & 25.0 & 79.5 & 32.8 & 20.2 & 43.9 & 44.2 & 51.9 \\
     HCL \cite{HCL} & \Checkmark & 80.9 & 34.9 & 76.7 & 6.6 & 0.2 & {36.1} & {20.1} & 28.2 & 79.1 & 83.1 & 55.6 & 25.6 & 78.8 & 32.7 & 24.1 & 32.7 & 43.5 & 50.2 \\

     LD \cite{LD-SFDA} & \Checkmark & 77.1 & 33.4 & 79.4 & 5.8 & 0.5 & 23.7 & 5.2 & 13.0 & 81.8 & 78.3 & 56.1 & 21.6 & 80.3 & 49.6 & 28.0 & 48.1 & 42.6 & 50.1 \\
     C-SFDA \cite{C-SFDA} & \Checkmark & 87.0 & 39.0 & 79.5 & 12.2 & 1.8 & 32.2 & 20.4 & 24.3 & 79.5 & 82.2 & 51.5 & 24.5 & 78.7 & 31.5 & 21.3 & 47.9 & 44.6 & 51.3 \\
     DTAC \cite{DTAC} & \Checkmark & 77.5 & 37.4 & 80.5 & 13.5 & 1.7 & 30.5 & 24.8 & 19.7 & 79.1 & 83.0 & 49.1 & 20.8 & 76.2 & 12.1 & 16.5 & 46.1 & 41.8 & 47.9 \\
     Cal-SFDA \cite{Cal-SFDA} & \Checkmark & 76.3 & 32.6 & 81.2 & 4.0 & 0.6 & 27.5 & 20.2 & 17.6 & 82.4 & 83.1 & 51.8 & 18.1 & 83.3 & 46.2 & 14.7 & 48.1 & 43.0 & 50.4 \\
     DT-ST \cite{DT-ST} & \Checkmark & 79.4 & 41.4 & 73.9 & 5.9 & 1.5 & 30.6 & 35.3 & 19.8 & 86.0 & 86.0 & 63.8 & 28.6 & 86.3 & 36.6 & 35.2 & 53.2 & 47.7 & 55.8 \\ 
     IAPC \cite{IAPC} & \Checkmark & 68.5 & 29.2 & 82.0 & 10.9 & 1.2 & 28.7 & 22.3 & 29.1 & 82.8 & 85.3 & 60.5 & 19.3 & 83.1 & 42.5 & 32.2 & 47.7 & 45.3 & 52.7 \\
     AUGCO \cite{AUGCO} & \Checkmark & 87.2 & 80.1 & 80.8 & 77.9 & 44.5 & 82.4 & 26.5 & 54.8 & 0.3 & 8.2 & 34.6 & 12.6 & 8.7 & 18.8 & 8.2 & 6.0 & 39.5 & 45.9 \\

     \midrule[0.7pt]

     \rowcolor{gray!40} \textbf{ATP (w/o TP)} & \Checkmark  & 89.7 & 44.6 & 80.1 & 3.3 & 0.3 & 28.4 & 8.1 & 7.7 & 81.1 & 85.2& 55.2 & 19.1 & 83.4 & 34.3 & 16.7 & 32.4 & 41.8 & 49.0 \\ 
     \rowcolor{gray!40} \textbf{ATP (w/o P)} & \Checkmark & 89.8 & 45.1 & {82.9} & 0.1 & 0.0 & 33.8 & 14.6 & 17.3 & 84.5 & 87.0 & 59.3 & 25.6 & 85.8 & 49.4 & 26.2 & 56.1  & 46.9 & 55.9 \\
     \rowcolor{gray!40} \textbf{ATP} & \Checkmark & 90.1 & {46.3} & 82.5 & 0.0 & 0.1 & 31.7 & 10.7 & 17.9 & {85.1} & 87.7 & {64.6} & {34.6} & 86.4 & {54.8} & 33.7 & 58.3 & 49.0 & 57.9 \\

     \midrule[0.7pt]
     \multicolumn{20}{c}{\textbf{P2T-Base}} \\
     \midrule[0.7pt]

     P2T \cite{P2T} & \Checkmark & 79.3 & 25.3 & 79.8 & 6.6 & 2.3 & 19.2 & 26.3 & 18.7 & 83.1 & 80.4 & 59.1 & 22.6 & 78.3 & 38.0 & 11.4 & 17.9 & 40.5 & 47.7 \\
     \rowcolor{gray!40} \textbf{ATP (w/o TP)} & \Checkmark & 89.9 & 27.8 & 82.9 & 5.6 & 9.3 & 29.2 & 26.5 & 19.9 & 84.2 & 84.9 & 57.7 & 21.6 & 79.8 & 47.0 & 14.9 & 18.7 & 43.7 & 50.4 \\
     \rowcolor{gray!40} \textbf{ATP (w/o P)} & \Checkmark & 92.2 & 47.9 & 83.5 & 10.3 & 11.5 & 37.3 & 31.7 & 25.6 & 84.4 & 86.8 & 61.6 & 25.0 & 87.6 & 43.7 & 30.2 & 36.4 & 49.7 & 56.7 \\
     \rowcolor{gray!40} \textbf{ATP} & \Checkmark & 90.2 & 48.2 & 85.5 & 10.3 & 13.3 & 37.9 & 31.1 & 25.9 & 86.7 & 89.7 & 67.6 & 35.7 & 89.8 & 52.5 & 33.3 & 38.4 & 52.3 & 59.6  \\ 

     \midrule[0.7pt]
     \multicolumn{20}{c}{\textbf{MiT-B5}} \\
     \midrule[0.7pt]      

     Source Only & - & 69.9 & 27.8 & 82.9 & 21.6 & 2.3 & 39.2 & 36.3 & 29.9 & 84.2 & 84.9 & 61.6 & 22.6 & 83.8 & 48.0 & 14.9 & 19.7 & 45.6 & 51.3 \\
     TransDA-B \cite{TransDA} & \XSolidBrush & 90.4 & 54.8 & 86.4 & 31.1 & 1.7 & 53.8 & 61.1 & 37.1 & 90.3 & 93.0 & 71.2 & 25.3 & 92.3 & 66.0 & 44.4 & 49.8 & 59.3 & 66.3 \\
     DAFormer \cite{DAFormer} & \XSolidBrush & 84.5 & 40.7 & 88.4 & 41.5 & 6.5 & 50.0 & 55.0 & 54.6 & 86.0 & 89.8 & 73.2 & 48.2 & 87.2 & 53.2 & 53.9 & 61.7 & 60.9 & 67.4 \\
     HRDA \cite{HRDA} & \XSolidBrush & 85.2 & 47.7 & 88.8 & 49.5 & 4.8 & 57.2 & 65.7 & 60.9 & 85.3 & 92.9 & 79.4 & 52.8 & 89.0 & 64.7 & 63.9 & 64.9 & 65.8 & 72.4 \\
     IDM \cite{IDM} & \XSolidBrush & 87.6 & 47.6 & 88.1 & 33.4 & 6.3 & 52.8 & 57.8 & 56.5 & 83.0 & 77.5 & 66.2 & 52.1 & 89.3 & 55.6 & 57.1 & 64.2 & 60.9 & 67.9 \\
    \midrule
     DAFormer \cite{DAFormer} & \Checkmark & 64.3 & 25.1 & 78.5 & 23.8 & 1.9 & 37.3 & 29.7 & 22.8 & 80.4 & 83.0 & 65.1 & 26.6 & 69.8 & 38.3 & 22.7 & 32.8 & 43.8 & 49.2 \\
     HRDA \cite{HRDA} & \Checkmark & 72.2 & 26.6 & 80.8 & 23.0 & 0.5 & 42.5 & 41.0 & 31.5 & 84.3 & 86.2 & 64.3 & 29.3 & 73.5 & 28.8 & 12.4 & 41.6 & 46.1 & 51.3 \\
     IDM \cite{IDM} & \Checkmark & 82.2 & 37.9 & 83.5 & 20.3 & 1.5 & 47.3 & 41.7 & 25.6 & 84.4 & 86.8 & 61.6 & 25.0 & 87.6 & 43.7 & 30.2 & 36.4 & 49.7 & 55.9 \\
     MISFIT \cite{MISFIT} & \Checkmark & 80.2 & 38.5 & 85.9 & 30.3 & 1.2 & 52.3 & 56.8 & 29.0 & 89.9 & 88.3 & 68.1 & 10.8 & 92.1 & 69.0 & 26.3 & 52.6 & 54.5 & 60.6 \\

      ELR \cite{ELR} & \Checkmark & - & - & - & - & - & - & - & - & - & - & - & - & - & - & - & - & 53.6 & 59.7 \\

      UVPT \cite{UVPT} & \Checkmark & - & - & - & - & - & - & - & - & - & - & - & - & - & - & - & - & 53.8 & 60.1 \\
      \midrule[0.7pt]

      \rowcolor{gray!40}\textbf{ATP (w/o TP)} & \Checkmark & 84.7 & 40.1 & 85.1 & 14.8 & 2.6 & 43.3 & 43.0 & 32.6 & 86.8 & 90.8 & 59.6 & 23.6 & 85.5 & 38.4 & 22.6 & 42.3 & 49.7 & 56.5 \\

      \rowcolor{gray!40}\textbf{ATP (w/o P)} & \Checkmark & 89.3 & 51.3 & 86.4 & 30.9 & 0.9 & 48.8 & 53.1 & 44.0 & 87.5 & 93.5 & 67.6 & 28.8 & 88.3 & 49.9 & 46.3 & 57.1 & 57.7 & 64.9 \\

      \rowcolor{gray!40}\textbf{ATP } & \Checkmark &  90.6 & 54.4 & 86.7 & 28.5 & 0.5 & 50.3 & 52.4 & 50.5 & 87.4 & 93.4 & 70.2 & 35.8 & 89.6 & 53.5 & 50.6 & 51.1 & 59.1 & 66.6 \\
     
     \bottomrule[1.0pt]
     \end{tabular}
     }
    \end{table*}
    
\subsection{Extension for Black-Box Source Model}

    \noindent In this section, we extend our method to a challenging but interesting black-box source model scenario \cite{DINE}, where only the target data and the corresponding predictions of the source model are available. Neither the raw source data nor the details about the source model is accessible during adaptation. It is a more practical scenario because we can utilize all kinds of server API as our source model. As no parameters are available, the source model can be treated as a black box, only providing the predictions of the target data. Due to no access to the trained source model parameters, the implicit feature alignment is impractical. Alternatively, a natural knowledge distillation (KD) \cite{KD} method is adopted to transfer the soft predictions from the source model to the target model. In this case, we pursue the knowledge distill to enforce the target model performing a similarity prediction with the source model. The knowledge distillation loss is formulated as follows:
    \begin{equation}
        \label{Equ:kd}
        \mathcal{L}_{kd} = \mathrm{E}_{x_t \in \mathcal{X}_t} \mathcal{D}_{kl}(f_s\circ g_s(x_t); f_t\circ g_t(x_t)),
        \end{equation}
    where $\mathcal{D}_{kl}$ denotes the Kullback-Leibler divergence loss. This step can transfer the knowledge from the source predictor to the target model. Then, to further fit the target model, we apply the proposed complementary self-training strategy and information propagation as the adaptation process. The details are illustrated in Figure \ref{Fig:BlackBox}. 


    In this section, we provide the algorithm of proposed \textbf{ATP} in Algorithm \ref{algorithm1} and \ref{algorithm}, which indicates our curriculum-style self-training strategy and the proposed standard post-processing for domain adaptation, respectively. Our curriculum-style self-training refers to first solving \textit{curriculum feature alignment} to encourage confident predictions and then \textit{curriculum complementary self-training}. The former enforces the target feature representations fitting to the source model from easy-to-hard adaptation, and the latter contains the self-training process utilizing the negative pseudo labels and the positive pseudo labels. During \textit{information propagation}, we propose standard post-processing for domain adaptation, which reduces intra-domain discrepancy in a semi-supervised learning manner.



\begin{table*}[!t]
     \centering
     \tabcolsep=2pt
     \footnotesize
     \caption{Semantic segmentation performance of Cityscapes$\rightarrow$ACDC trained on different ResNet-101 and MiT-B5 backbones. ``SF" denotes the source data-free setting. }\label{Tab:CS2ACDC}
     \resizebox{\textwidth}{!}{
     \begin{tabular}{lccccccccccccccccccccc}
       \toprule[1.0pt]
        {Method} & {SF} & \rotatebox{90}{road} & \rotatebox{90}{side.} & \rotatebox{90}{build.} & \rotatebox{90}{wall}   & \rotatebox{90}{fence} & \rotatebox{90}{pole} & \rotatebox{90}{light}  & \rotatebox{90}{sign}  & \rotatebox{90}{vege.} & \rotatebox{90}{terr.}  & \rotatebox{90}{sky}   & \rotatebox{90}{person} & \rotatebox{90}{rider}  & \rotatebox{90}{car}   & \rotatebox{90}{truck} & \rotatebox{90}{bus}    & \rotatebox{90}{train} & \rotatebox{90}{motor.} & \rotatebox{90}{bike}   & \rotatebox{90}{\textbf{mIoU}} \\

     \toprule[1.0pt]
     \multicolumn{22}{c}{\textbf{ResNet-101}} \\
     \midrule[0.7pt]
      Source Model & - & 76.6 & 40.5 & 56.0 & 12.0 & 27.3 & 35.6 & 40.2 & 45.6 & 69.8 & 38.2 & 76.2 & 21.3 & 12.4 & 65.6 & 25.2 & 29.2 & 28.1 & 15.2 & 34.6 & 39.5 \\
      AdaptSeg \cite{AdaptSeg} & \XSolidBrush & 69.4 & 34.0 & 52.8 & 13.5 & 18.0 & 4.3 & 14.9 & 9.7 & 64.0 & 23.1 & 38.2 & 38.6 & 20.1 & 59.3 & 35.6 & 30.6 & 53.9 & 19.8 & 33.9 & 33.4 \\
      ADVENT \cite{ADVENT} & \XSolidBrush & 72.9 & 14.3 & 40.5 & 16.6 & 21.2 & 9.3 & 17.4 & 21.2 & 63.8 & 23.8 & 18.3 & 32.6 & 19.5 & 69.5 & 36.2 & 34.5 & 46.2 & 26.9 & 36.1 & 32.7 \\
      FDA \cite{FDA} & \XSolidBrush & 73.2 & 34.7 & 59.0 & 24.8 & 29.5 & 28.6 & 43.3 & 44.9 & 70.1 & 28.2 & 54.7 & 47.0 & 28.5 & 74.6 & 44.8 & 52.3 & 63.3 & 28.3 & 39.5 & 45.7 \\
      DACS \cite{DACS} & \XSolidBrush & 58.5 & 34.7 & 76.4 & 20.9 & 22.6 & 31.7 & 32.7 & 46.8 & 58.7 & 39.0 & 36.3 & 43.7 & 20.5 & 72.3 & 39.6 & 34.8 & 51.1 & 24.6 & 38.2 & 41.2 \\
      VBLC \cite{VBLC} & \XSolidBrush & 49.6 & 39.3 & 79.4 & 35.8 & 29.5 & 42.6 & 57.2 & 57.5 & 69.1 & 42.7 & 39.8 & 54.5 & 29.3 & 77.8 & 43.0 & 36.2 & 32.7 & 38.7 & 53.4 & 47.8 \\
      \midrule[0.7pt]
      HCL \cite{HCL} & \Checkmark &  80.5 & 42.9 & 57.6 & 14.7 & 29.4 & 40.3 & 49.0 & 51.1 & 72.4 & 35.6 & 78.3 & 39.7 & 31.8 & 76.0 & 35.4 & 42.7 & 42.5 & 25.7 & 43.0 & 46.8 \\
      URMA \cite{UncertaintyReducing} & \Checkmark & 85.4 & 52.9 & 62.9 & 20.4 & 34.4 & 39.9 & 36.7 & 43.9 & 74.9 & 46.9 & 85.1 & 27.2 & 22.4 & 76.0 & 40.5 & 41.5 & 38.9 & 20.6 & 46.2 & 47.2 \\
      SimT \cite{SimT} & \Checkmark &  83.5 & 52.7 & 60.7 & 19.6 & 33.7 & 42.0 & 43.1 & 47.4 & 75.0 & 42.5 & 85.8 & 39.8 & 19.6 & 76.9 & 39.6 & 42.7 & 41.1 & 24.0 & 43.1 & 48.0 \\
      CMA \cite{CMA} & \Checkmark & 83.1 & 52.7 & 65.4 & 18.7 & 30.5 & 44.5 & 56.3 & 53.9 & 76.7 & 39.7 & 79.0 & 54.2 & 31.2 & 76.7 & 40.2 & 39.3 & 47.4 & 29.8 & 38.6 & 50.4 \\

      \midrule[0.7pt]
      \rowcolor{gray!40}\textbf{ATP (w/o TP)} & \Checkmark & 
      67.4 & 31.2 & 64.0 & 33.5 & 29.8 & 36.6 & 48.7 & 55.8 & 84.6 & 29.9 & 68.2 & 31.0 & 16.7 & 69.5 & 36.0 & 23.0 & 44.4 & 26.7 & 22.5 & 43.1 \\

      \rowcolor{gray!40}\textbf{ATP (w/o P)} & \Checkmark & 
      72.0 & 42.3 & 69.8 & 41.4 & 29.5 & 40.4 & 53.4 & 60.2 & 87.0 & 33.0 & 77.1 & 54.7 & 20.3 & 74.5 & 27.6 & 16.2 & 49.1 & 32.5 & 39.5 & 48.4 \\

      \rowcolor{gray!40}\textbf{ATP } & \Checkmark & 
      76.2 & 47.3 & 71.4 & 42.7 & 31.4 & 44.2 & 55.4 & 62.0 & 89.0 & 34.7 & 79.1 & 49.9 & 16.6 & 77.5 & 30.0 & 19.7 & 47.7 & 44.0 & 39.4 & 50.5 \\

      \midrule[0.7pt]
      \multicolumn{22}{c}{\textbf{MiT-B5}} \\
      \midrule[0.7pt]
      TENT \cite{TENT} & \Checkmark & 85.3 & 50.2 & 85.4 & 45.4 & 32.7 & 50.4 & 59.4 &	66.1 &	86.4 & 45.7 & 97.5 &	57.9 & 53.8 & 84.7 &	51.0 & 66.9 & 72.4 & 40.2 &	50.1 & 62.2 \\
      CoTTA \cite{CoTTA} & \Checkmark & 85.7	& 50.9 & 85.9 & 45.9 & 33.6 & 54.8 & 62.3 & 69.9 & 87.1 & 45.7 & 97.7 &	63.3 & 59.4 & 85.1 & 52.8 & 68.0 & 74.1 & 44.9 & 55.1 & 64.4 \\
      DePT \cite{DePT} & \Checkmark & 85.0 &	50.6 & 85.5 & 45.7 & 33.2 & 53.9 & 61.6 & 69.4 & 86.7 & 45.5 & 97.4 & 62.6 & 59.2 &	85.1 & 52.5 & 68.0 & 73.7 & 44.3 & 54.3 & 63.9 \\
      VDP \cite{VDP} & \Checkmark & 85.7	& 50.9 & 85.9 & 45.9 & 33.6 & 54.8 & 62.2 &	69.9 & 87.0 & 45.7 & 97.6	& 63.3 & 59.2 &	85.1 & 52.8 & 68.1 & 74.1 & 44.8 & 54.9 &	64.3 \\
      IDM \cite{IDM} & \Checkmark & 88.8	& 63.2 & 85.8 &	45.5 & 30.3 & 42.1 & 69.7 &	62.7 &	87.4 & 51.7 & 96.5 &	61.8 & 29.3 & 86.5 &	80.5 & 68.9 & 70.1 & 57.0 &	52.8 & 64.8 \\

      \midrule[0.7pt]

      \rowcolor{gray!40} \textbf{ATP (w/o TP)} & \Checkmark & 79.7 & 44.1 & 72.2 & 32.4 & 32.6 & 50.4 & 67.8 & 61.7 & 88.8 & 44.2 & 83.3 & 55.9 &	22.1 & 78.9 & 49.2 &	59.4 & 71.4 & 34.5 & 39.1 &	56.2 \\
      \rowcolor{gray!40} \textbf{ATP (w/o P)} & \Checkmark & 88.4 &	62.8 & 84.2 & 40.7 & 30.2 & 51.7 & 72.3 & 65.0 & 89.9 & 45.8 & 94.8 & 66.1 & 28.6 & 85.4 &	63.6 & 72.5 & 82.5 & 44.8 & 54.8 & 64.4 \\
      \rowcolor{gray!40} \textbf{ATP} & \Checkmark & 88.4 &	63.2 & 87.4 & 50.0 & 32.5 & 53.5 & 71.8 & 66.9 &	90.3 & 42.7 & 97.0 & 66.3 & 32.1 & 85.8 &	63.9 & 70.5 & 83.2 & 46.1 & 53.8 & 65.6 \\
      \bottomrule[1.0pt]
      \end{tabular}
      }
    \end{table*}

    
    \begin{table}[!tbp]
        \centering
        \footnotesize
        \tabcolsep=6pt
        \caption{Semantic segmentation performance of Cityscapes $\rightarrow$ Cross-City domain adaptation. ``SF" denotes the source data-free setting.}\label{Tab:CS2CC}
        \begin{tabular}{ccccccc}
        \toprule
        Method & SF & Rome & Rio & Tokyo & Taipei & Mean \\
        \midrule[0.7pt]
        ADVENT \cite{ADVENT} & \XSolidBrush & 47.3 & 46.8 & 45.5 & 45.1 & 45.3 \\
        AdaptSegNet \cite{AdaptSeg} & \XSolidBrush & 48.0 & 47.8 & 46.2 & 45.1 & 46.8 \\
        Cross-City \cite{Cross-City} & \XSolidBrush & 42.9 & 42.5 & 42.8 & 39.6 & 42.0 \\
        MaxSquare \cite{MaxSquare} & \XSolidBrush & 54.5 & 53.3 & 50.5 & 50.6 & 52.2 \\
        \midrule[0.7pt] 
        SFDA \cite{SFDA} & \Checkmark & 48.3 & 49.0 & 46.4 & 47.2 & 47.7 \\
        URMA \cite{UncertaintyReducing} & \Checkmark & 53.8 & 53.5 & 49.8 & 50.1 & 51.8 \\
        \rowcolor{gray!40} \textbf{ATP (w/o TP)} & \Checkmark & 48.2 & 48.7 & 45.3 & 43.6 & 46.5 \\
        \rowcolor{gray!40} \textbf{ATP (w/o P)} & \Checkmark & 53.8 & 55.3 & 48.6 & 49.6 & 51.8 \\
        \rowcolor{gray!40} \textbf{ATP} & \Checkmark & 55.9 & 57.8 & 50.5 & 51.1 & 53.8  \\
        \bottomrule
        \end{tabular}
    \end{table}


    \begin{table*}[!tbp]
     \centering
     \tabcolsep=4pt
     \caption{Results of \emph{black-box} source model on GTA5 $\rightarrow$ Cityscapes domain adaptation. ``SF" denotes the source data-free setting. \textbf{\textcolor{gray}{Gray}} row group denotes results of our black-box source model.}\label{Tab:GTA5}
     \resizebox{\textwidth}{!}{
     \begin{tabular}{lccccccccccccccccccccc} 
       \toprule[1.0pt]
        {Method} & SF & \rotatebox{90}{road} & \rotatebox{90}{side.} & \rotatebox{90}{build.} & \rotatebox{90}{wall}   & \rotatebox{90}{fence} & \rotatebox{90}{pole} & \rotatebox{90}{light}  & \rotatebox{90}{sign}  & \rotatebox{90}{vege.} & \rotatebox{90}{terr.}  & \rotatebox{90}{sky}   & \rotatebox{90}{person} & \rotatebox{90}{rider}  & \rotatebox{90}{car}   & \rotatebox{90}{truck} & \rotatebox{90}{bus}    & \rotatebox{90}{train} & \rotatebox{90}{motor.} & \rotatebox{90}{bike}   & \rotatebox{90}{\textbf{mIoU}} \\

      \midrule[1.0pt]

      AdaptSeg \cite{AdaptSeg} & \XSolidBrush & 86.5 & 36.0 & 79.9 & 23.4 & 23.3 & 23.9 & 35.2 & 14.8 & 83.4 & 33.3 & 75.6 & 58.5 & 27.6 & 73.7 & 32.5 & 35.4 & 3.9 & 30.1 & 28.1 & 42.4 \\
      ADVENT \cite{ADVENT} & \XSolidBrush & 89.4 & 33.1 & 81.0 & 26.6 & 26.8 & 27.2 & 33.5 & 24.7 & 83.9 & 36.7 & 78.8 & 58.7 & 30.5 & 84.8 & 38.5 & 44.5 & 1.7 & 31.6 & 32.4 & 45.5 \\
      CBST \cite{CBST} & \XSolidBrush & 91.8 & 53.5 & 80.5 & 32.7 & 21.0 & 34.0 & 28.9 & 20.4 & 83.9 & 34.2 & 80.9 & 53.1 & 24.0 & 82.7 & 30.3 & 35.9 & 16.0 & 25.9 & 42.8 & 45.9 \\
      MaxSquare \cite{MaxSquare} & \XSolidBrush & 89.4 & 43.0 & 82.1 & 30.5 & 21.3 & 30.3 & 34.7 & 24.0 & 85.3 & 39.4 & 78.2 & 63.0 & 22.9 & 84.6 & 36.4 & 43.0 & 5.5 & 34.7 & 33.5 & 46.4 \\
      PLCA \cite{PLCA} & \XSolidBrush & 84.0 & 30.4 & 82.4 & 35.3 & 24.8 & 32.2 & 36.8 & 24.5 & 85.5 & 37.2 & 78.6 & 66.9 & 32.8 & 85.5 & 40.4 & 48.0 & 8.8 & 29.8 & 41.8 & 47.7 \\
      
      \midrule[0.7pt]
      
      SFDA \cite{SFDA} & \Checkmark & 84.2 & 39.2 & 82.7 & 27.5 & 22.1 & 25.9 & 31.1 & 21.9 & 82.4 & 30.5 & 85.3 & 58.7 & 22.1 & 80.0 & 33.1 & 31.5 & 3.6 & 27.8 & 30.6 & 43.2 \\
      URMA \cite{UncertaintyReducing} & \Checkmark & 92.3 & 55.2 & 81.6 & 30.8 & 18.8 & 37.1 & 17.7 & 12.1 & 84.2 & 35.9 & 83.8 & 57.7 & 24.1 & 81.7 & 27.5 & 44.3 & 6.9 & 24.1 & 40.4 & 45.1 \\
      S4T \cite{S4T} & \Checkmark & 89.7 & 39.2 & 84.4 & 25.7 & 29.0 & 39.5 & 45.1 & 36.8 & 86.8 & 41.8 & 79.3 & 61.2 & 26.7 & 85.0 & 19.3 & 28.2 & 5.3 & 11.8 & 9.3 & 44.8 \\ 
      HCL \cite{HCL} & \Checkmark & 93.3 & 58.0 & 81.9 & 23.8 & 24.5 & 24.9 & 8.5 & 31.4 & 84.2 & 37.4 & 84.6 & 57.4 & 24.2 & 84.1 & 29.1 & 39.9 & 0.0 & 33.1 & 47.5 & 45.7 \\
      DINE \cite{DINE} & \Checkmark & 88.2 & 44.2 & 83.5 & 14.1 & 32.4 & 23.5 & 24.6 & 36.8 & 85.4 & 38.3 & 85.3 & 59.8 & 27.4 & 84.7 & 30.1 & 42.2 & 0.0 & 42.7 & 45.3 & 46.7 \\
    
      \midrule[0.7pt]           


       Baseline & - & 75.5 & 25.4 & 80.8 & 20.2 & 25.0 & 10.9 & 17.5 & 0.0 & 84.4 & 16.0 & 83.5 & 58.7 & 0.0 & 80.5 & 34.9 & 46.1 & 0.0 & 31.9 & 42.0 & 38.6  \\

       \rowcolor{gray!40} \textbf{ATP} (w/o P) & \Checkmark & 74.8 & 23.6 & 78.9 & 27.4 & 17.5 & 24.1 & 34.0 & 24.9 & 82.6 & 25.7 & 74.3 & 58.6 & 31.7 & 71.5 & 31.4 & 6.9 & 0.0 & 34.8 & 45.4 & 40.4 \\
       \rowcolor{gray!40} \textbf{ATP} & \Checkmark & 83.3 & 24.2 & 80.3 & 29.7 & 22.4 & 26.3 & 33.0 & 25.5 & 84.7 & 35.7 & 76.6 & 61.9 & 34.0 & 81.7 & 32.1 & 17.3 & 0.0 & 42.1 & 53.4 & 44.4 \\
       \rowcolor{gray!40} \textbf{ATP} (aug) & \Checkmark & 83.6 & 25.8 & 81.9 & 30.2 & 25.2 & 27.9 & 36.2 & 28.7 & 84.8 & 34.4 & 77.5 & 62.2 & 35.7 & 81.5 & 32.3 & 16.8 & 0.0 & 41.7 & 53.5 & 45.3 \\

       \bottomrule[1.0pt]
      \end{tabular}
      }
    \end{table*}

\section{Experiments}\label{Sec:Exp}

\subsection{Datasets}

    \noindent We demonstrate the efficacy of the proposed method on the standard adaptation tasks of GTA5 \cite{GTA5}$\rightarrow$Cityscapes \cite{Cityscapes}, SYNTHIA \cite{SYNTHIA}$\rightarrow$Cityscapes, Cityscapes$\rightarrow$NTHU Cross-City \cite{Cross-City} and Cityscapes$\rightarrow$ACDC \cite{ACDC}. The synthetic dataset \textbf{GTA5} contains 24,966 annotated images with a resolution of 1914$\times$1052, token from the famous game Grand Theft Auto. The ground truth is generated by the game rendering itself. \textbf{SYNTHIA} is another synthetic dataset, which contains 9,400 fully annotated images with a resolution of 1280 $\times$ 760. \textbf{Cityscapes} consists of 2,975 annotated training images and 500 validation images with a resolution of 2048 $\times$ 1024. \textbf{NTHU Cross-City} dataset has been recorded in four cities: Rome, Rio, Tokyo, and Taipei. Each city set has 3,200 unlabeled training images and 100 testing images with a resolution of 2048 $\times$ 1024. The Adverse Conditions Dataset (\textbf{ACDC}) \cite{ACDC} contains four different adverse visual conditions: Fog, Night, Rain, and Snow. Images in ACDC share the same semantic classes with Cityscapes. Therefore, we also conduct adaptation experiments on Cityscapes to ACDC. We employ the mean Intersection over Union (mIoU) as the evaluation metric, which is widely adopted in semantic segmentation tasks.

    Moreover, we also evaluate our method on image recognition tasks, including \textbf{VisDA-C} \cite{VisDA-C} and \textbf{OfficeHome} \cite{OfficeHome} datasets. Specifically, VisDA-C is a challenging large-scale dataset for adaptation algorithms from synthetic-to-real tasks. There are 152,397 synthetic images in the source domain and 55,388 real-world images in the target domain. OfficeHome dataset consists of four different domains: Artistic (Ar), Clipart (Cl), Product (Pr), and Real-World (Rw). Each domain contains 65 categories.

\subsection{Implementation Details}
\label{details}

    \noindent We conduct experiments based on CNN-based architectures with ResNet-101 \cite{ResNet}, P2T-Base \cite{P2T}, and Transformer-based architectures with MiT-B5 \cite{SegFormer} as backbones, which is the same as previous works \cite{AdaptSeg,CLAN,ADVENT,UncerDA,DAFormer}. We first pre-train the segmentation network on the source domain for 80k iterations to obtain a high-quality source model. Then it is considered as the initialization model. During training, we follow previous works \cite{AdaptSeg,CLAN,ADVENT} using Stochastic Gradient Descent (SGD) optimizer with the learning rate $2.5 \times 10^{-4}$, momentum 0.9, and weight decay $5 \times 10^{-4}$. We schedule the learning rate using ``poly" policy: the learning rate is multiplied by $(1-\frac{iter}{max\_iter})^{0.9}$. In the entropy minimization stage, we fix the classifier as mentioned above. In the other stages, the classifier is optimized with the learning rate $2.5 \times 10^{-3}$. Concerning parameters, we follow the unsupervised hyper-parameters selection method \cite{SFD} to tune the following parameters. We first freeze $\gamma$ and choose $\alpha=0.002$ from the range of [0.001, 0.002, 0.003], and then the value of $\gamma$ is selected as 3 from the range of [0,1,2,3,4,5]. For the self-training process, we assign the threshold $\lambda_c$ according to the category-level top $65\%$ target predictions to generate pseudo-labels, and the threshold of negative pseudo labels $\lambda_{neg}$ is assigned as 0.05. We update pseudo labels epoch by epoch, and the maximum number of epochs is empirically set as 10. For the semi-supervised training process, we assign the target images with average entropy ranked top $50\%$ as the easy group, and the other images are defined as the hard group. Generally, we randomly run our methods three times with different random seeds and report the average accuracy.

\subsection{Results}

    \noindent In this section, we show the performance of proposed \textbf{ATP} on GTA5$\rightarrow$Cityscapes in Table \ref{Tab:GTA2CS}, SYNTHIA$\rightarrow$Cityscapes in Table \ref{Tab:SYN2CS}, and Cityscapes$\rightarrow$ACDC in Table \ref{Tab:CS2ACDC} based on different backbones. ATP (w/o TP), ATP (w/o P), and ATP denote results of different training stages: curriculum feature alignment, complementary curriculum self-training, and information propagation, respectively. In addition, we compare our method with previous state-of-the-art works, including traditional source data-dependent methods~\cite{AdaptSeg,ADVENT,CBST,CAG,FDA,MaxSquare,PLCA,SIM,DAFormer,HRDA} and source data-free methods~\cite{SFDA,UncertaintyReducing,S4T,SFUDA,HCL}. The utility of source data can directly align distributions of source and target domains. Thus, conventional source data-dependent approaches obtain higher performance than source data-free adaptation methods reasonably. Notice that our method \textbf{ATP} performing domain adaptation without source data achieves comparable results or even better than source data-dependent methods. Furthermore, concerning source data-free scenarios, our method outperforms previous works \cite{SFDA,UncertaintyReducing,HCL,S4T,SFUDA} with a large margin. Compared to recent source data-free methods, we can observe that the proposed ATP still performs better than them, including Cal-SFDA \cite{Cal-SFDA}, C-SFDA \cite{C-SFDA}, DT-ST \cite{DT-ST}, and so on.

    Specifically, for the tasks of GTA5$\rightarrow$Cityscapes and SYNTHIA$\rightarrow$Cityscapes, as shown in Table \ref{Tab:GTA2CS} and \ref{Tab:SYN2CS}, our method arrives 52.6\% and 57.9\% scores based on ResNet-101, respectively, which offers a large margin performance gain compared to the non-adaptation baseline. Moreover, the proposed method surpasses four existing source data-free methods SFDA \cite{SFDA}, URMA \cite{UncertaintyReducing}, SFUDA \cite{SFUDA}, and HCL \cite{HCL}, with 9.4\%, 7.5\%, 3.2\%, and 4.5\% mIoU scores on GTA5$\rightarrow$Cityscapes, and 12.0\%, 12.9\%, 6.0\%, and 7.7\% mIoU scores on SYNTHIA$\rightarrow$Cityscapes, respectively. Compared with the traditional domain adaptation approaches that utilize the source data, our method achieves comparable or superior results. These results verify the effectiveness of our method. Besides, considering the transformer-based backbone, our method also provides a large margin improvement compared to existing state-of-the-art methods \cite{DAFormer,IDM,HRDA}, with the performance of 64.0\% and 66.6\% mIoU for GTA5$\rightarrow$Cityscapes and SYNTHIA$\rightarrow$Cityscapes, respectively. Moreover, we also conduct experiments based on P2T-Base \cite{P2T} backbone on GTA5/SYNTHIA $\rightarrow$Cityscapes tasks. From the results, we can observe that the proposed ATP achieves a margin improvement compared to the baseline method P2T \cite{P2T} with performing 55.8\% and 59.6\% mIoU.

\begin{table*}[!tbp]
     \centering
     \tabcolsep=7pt
     \footnotesize
     \caption{Classification performance under SFDA settings on VisDA dataset (ResNet101 backbone). We report the accuracy of each category and take the average (Avg.) over them.}\label{Tab:VisDA}
     \resizebox{\textwidth}{!}{
     \begin{tabular}{lcccccccccccccc}
     \toprule[1.0pt]
     Method & SF & plane & bike & bus & car & horse & knife & mcycle & person & plant & sktbrd & train & truck & Avg. \\

     \toprule[1.0pt]
     Source Only & - & 57.2 & 11.1 & 42.4 & 66.9 & 55.0 & 4.4 & 81.1 & 27.3 & 57.9 & 29.4 & 86.7 & 5.8 & 43.8 \\
     MCC \cite{MCC} & \XSolidBrush & 88.7 & 80.3 & 80.5 & 71.5 & 90.1 & 93.2 & 85.0 & 71.6 & 89.4 & 73.8 & 85.0 & 36.9 & 78.8 \\
     STAR \cite{STAR} & \XSolidBrush & 95.0 & 84.0 & 84.6 & 73.0 & 91.6 & 91.8 & 85.9 & 78.4 & 94.4 & 84.7 & 87.0 & 42.2 & 82.7 \\
     RWOT \cite{RWOT} & \XSolidBrush & 95.1 & 80.3 & 83.7 & 90.0 & 92.4 & 68.0 & 92.5 & 82.2 & 87.9 & 78.4 & 90.4 & 68.2 & 84.0 \\
     SE \cite{SE} & \XSolidBrush & 95.9 & 87.4 & 85.2 & 58.6 & 96.2 & 95.7 & 90.6 & 80.0 & 94.8 & 90.8 & 88.4 & 47.9 & 84.3 \\
        
     \midrule[0.7pt]

     SHOT \cite{SHOT} & \Checkmark & 94.3 & 88.5 & 80.1 & 57.3 & 93.1 & 94.9 & 80.7 & 80.3 & 91.5 & 89.1 & 86.3 & 58.2 & 82.9 \\
     A$^{2}$Net \cite{A2Net} & \Checkmark & 94.0 & 87.8 & 85.6 & 66.8 & 93.7 & 95.1 & 85.8 & 81.2 & 91.6 & 88.2 & 86.5 & 56.0 & 84.3 \\
     SFDA-DE \cite{SFDA-DE} & \Checkmark & 95.3 & 91.2 & 77.5 & 72.1 & 95.7 & 97.8 & 85.5 & 86.1 & 95.5 & 93.0 & 86.3 & 61.6 & 86.5 \\
     AdaCon \cite{AdaCon} & \Checkmark & 97.0 & 84.7 & 84.0 & 77.3 & 96.7 & 93.8 & 91.9 & 84.8 & 94.3 & 93.1 & 94.1 & 49.7 & 86.8 \\
     C-SFDA \cite{C-SFDA} & \Checkmark & 97.6 & 88.8 & 86.1 & 72.2 & 97.2 & 94.4 & 92.1 & 84.7 & 93.0 & 90.7 & 93.1 & 63.5 & 87.8 \\
     NRC ++ \cite{NRC} & \Checkmark & 97.4 & 91.9 & 88.2 & 83.2 & 97.3 & 96.2 & 90.2 & 81.1 & 96.3 & 94.3 & 91.4 & 49.6 & 88.1 \\
     PLUE \cite{PLUE} & \Checkmark & 97.3 & 96.2 & 90.5 & 91.8 & 90.0 & 94.2 & 87.4 & 87.7 & 97.0 & 84.3 & 93.0 & 81.0 & 90.0 \\

     \midrule[0.7pt]

     \rowcolor{gray!40} \textbf{ATP (w/o TP)} & \Checkmark & 98.0 & 99.1 & 48.3 & 56.5 & 95.3 & 99.9 & 68.3 & 95.9 & 95.9 & 73.9 & 69.6 & 74.5 & 81.3 \\ 
     \rowcolor{gray!40} \textbf{ATP (w/o P)} & \Checkmark & 94.7 & 84.0 & 87.5 & 71.9 & 95.3 & 87.3 & 92.0 & 83.7 & 92.8 & 91.1 & 89.6 & 48.7 & 84.9 \\
     \rowcolor{gray!40} \textbf{ATP} & \Checkmark & 97.6 & 91.8 & 88.7 & 73.1 & 97.6 & 92.9 & 92.0 & 95.7 & 93.4 & 89.0 & 87.9 & 71.3 & 89.3  \\

      \bottomrule[1.0pt]
      \end{tabular}
      }
    \end{table*}

\begin{table*}[!tbp]
     \centering
     \tabcolsep=2pt
     \footnotesize
     \caption{Classification performance under SFDA settings on Office-Home dataset (ResNet50 backbone). We report Top-1 accuracy on 12 domain shifts and take the average (Avg.) over them.}\label{Tab:OfficeHome}
     \resizebox{\textwidth}{!}{
     \begin{tabular}{lcccccccccccccc}
     \toprule[1.0pt]
     Method & SF & Ar $\rightarrow$ Cl & Ar $\rightarrow$ Pr & Ar $\rightarrow$ Rw & Cl $\rightarrow$ Ar & Cl $\rightarrow$ Pr & Cl $\rightarrow$ Rw & Pr $\rightarrow$ Ar & Pr $\rightarrow$ Cl & Pr $\rightarrow$ Rw & Rw $\rightarrow$ Ar & Rw $\rightarrow$ Cl & Rw $\rightarrow$ Pr & Avg. \\

     \toprule[1.0pt]
     RSDA \cite{RSDA} & \XSolidBrush & 53.2 & 77.7 & 81.3 & 66.4 & 74.0 & 76.5 & 67.9 & 53.0 & 82.0 & 75.8 & 57.8 & 85.4 & 70.9 \\
     TSA \cite{TSA} & \XSolidBrush & 57.6 & 75.8 & 80.7 & 64.3 & 76.3 & 75.1 & 66.7 & 55.7 & 81.2 & 75.7 & 61.9 & 83.8 & 71.2 \\
     SRDC \cite{SRDC} & \XSolidBrush & 52.3 & 76.3 & 81.0 & 69.5 & 76.2 & 78.0 & 68.7 & 53.8 & 81.7 & 76.3 & 57.1 & 85.0 & 71.3 \\
     FixBi \cite{FixBi} & \XSolidBrush & 58.1 & 77.3 & 80.4 & 67.7 & 79.5 & 78.1 & 65.8 & 57.9 & 81.7 & 76.4 & 62.9 & 86.7 & 72.7 \\
        
     \midrule[0.7pt]

     G-SFDA \cite{G-SFDA} & \Checkmark &  57.9 & 78.6 & 81.0 & 66.7 & 77.2 & 77.2 & 65.6 & 56.0 & 82.2 & 72.0 & 57.8 & 83.4 & 71.3 \\
     SHOT \cite{SHOT} & \Checkmark & 57.1 & 78.1 & 81.5 & 68.0 & 78.2 & 78.1 & 67.4 & 54.9 & 82.2 & 73.3 & 58.8 & 84.3 & 71.8 \\
     HCL \cite{HCL} & \Checkmark & 64.0 & 78.6 & 82.4 & 64.5 & 73.1 & 80.1 & 64.8 & 59.8 & 75.3 & 78.1 & 69.3 & 81.5 & 72.6 \\
     A$^{2}$Net \cite{A2Net} & \Checkmark & 58.4 & 79.0 & 82.4 & 67.5 & 79.3 & 78.9 & 68.0 & 56.2 & 82.9 & 74.1 & 60.5 & 85.0 & 72.8 \\
     SFDA-DE \cite{SFDA-DE} & \Checkmark & 59.7 & 79.5 & 82.4 & 69.7 & 78.6 & 79.2 & 66.1 & 57.2 & 82.6 & 73.9 & 60.8 & 85.5 & 72.9 \\
     C-SFDA \cite{C-SFDA} & \Checkmark & 60.3 & 80.2 & 82.9 & 69.3 & 80.1 & 78.8 & 67.3 & 58.1 & 83.4 & 73.6 & 61.3 & 86.3 & 73.5 \\
     NRC++ \cite{NRC} & \Checkmark & 57.8 & 80.4 & 81.6 & 69.0 & 80.3 & 79.5 & 65.6 & 57.0 & 83.2 & 72.3 & 59.6 & 85.7 & 72.5 \\

     \midrule[0.7pt]

     \rowcolor{gray!40} \textbf{ATP (w/o TP)} & \Checkmark & 58.5 & 75.8 & 79.9 & 67.0 & 73.4 & 74.3 & 65.7 & 54.7 & 80.6 & 74.1 & 59.6 & 82.9 & 70.5 \\ 
     \rowcolor{gray!40} \textbf{ATP (w/o P)} & \Checkmark & 58.8 & 76.8 & 80.8 & 67.9 & 81.2 & 74.9 & 65.7 & 55.2 & 80.8 & 73.1 & 61.4 & 83.7 & 71.7 \\
     \rowcolor{gray!40} \textbf{ATP} & \Checkmark & 59.3 & 78.5 & 82.6 & 69.4 & 82.4 & 77.7 & 68.0 & 56.3 & 82.3 & 74.3 & 62.3 & 84.5 & 73.1 \\

      \bottomrule[1.0pt]
      \end{tabular}
      }
    \end{table*}

    As shown in Table \ref{Tab:CS2ACDC}, we extend the proposed ATP to a more challenging task, from real Cityscapes to adverse conditions weather images ACDC, containing images in fog, night, rain, and snow. In comparison, our method performs a stable improvement benefiting from the curriculum feature alignment, curriculum complementary self-training, and information propagation. Specifically, we perform experiments based on different backbones, including ResNei-101 and MiT-B5. We modify existing methods for fair comparison on the source data-free scenario, including HCL \cite{HCL}, URMA \cite{UncertaintyReducing}, SimT \cite{SimT}, CMA \cite{CMA}, DePT \cite{DePT}, VDP \cite{VDP}, and IDM \cite{IDM}. The results reveal that our method performs state-of-the-art performance in these tasks.

    In Table \ref{Tab:CS2CC}, we show results on Cityscapes$\rightarrow$Cross-City task that has a minor domain shift attribute to both source and target data obtained from real-world datasets. We also compare with conventional source data-dependent domain adaptation approaches \cite{ADVENT,AdaptSeg,Cross-City,MaxSquare} and source data-free approaches \cite{SFDA,UncertaintyReducing} that provide adaptation results in the original paper. From the results, we can observe that our method \textbf{ATP} achieves the best performance in four different adaptation scenarios, including from Cityscapes to Rome, Rio, Tokyo, and Taipei, respectively. It demonstrates the effectiveness of our method for minor domain shift adaptation.

\subsection{Results on Image Recognition}

    To verify the effectiveness and generalization, we also apply the proposed method to image recognition under source-free domain adaptation settings, including the adaptation on VisDA-C and OfficeHome datasets. The corresponding results are shown in Table \ref{Tab:VisDA} and \ref{Tab:OfficeHome}. Following previous works, we provide the performance of VisDA-C and OfficeHome datasets based on the ResNet-101 and ResNet-10 backbones, respectively. From the results, we can observe that our method performs comparable results to the recent state-of-the-art works, exampling as C-SFDA \cite{C-SFDA}, PLUE \cite{PLUE}, and NRC++ \cite{NRC}. These results demonstrate the effectiveness and generalizability of the proposed method.

\subsection{Extension for Black-box Source Model}
    \noindent In Table \ref{Tab:GTA5}, we provide the results of the black-box source model case where only the source model's predictions are available.
    This situation is more challenging but interesting in practice because it does not require any source model details but the network outputs. Due to no access to trained source model parameters, we use knowledge distillation to enforce the target model to learn similarity predictions with the source model, which is considered as the baseline model with 38.6\% mIoU performance. After that, we apply the proposed complementary self-training strategy and information propagation to tackle the black-box source model scenario. ``\textbf{ATP} (w/o P)" and ``\textbf{ATP}" represent our black-box source model on the complementary self-training stage and information propagation stage, respectively. Furthermore, we apply data augmentation techniques like ``Color Jitter" and ``Gaussian Blur" to enhance feature representations. ``\textbf{ATP} (aug)" denotes the corresponding results. In Table \ref{Tab:GTA5}, we compare the black-box source model with conventional domain adaptation and source data-free domain adaptation approaches. 
    From the results, we can observe that our method is suitable for the black-box source model situation with a performance of 45.3\%, achieving comparable results to source data-dependent and source data-free adaptation methods.

\subsection{Ablation Study}

    \noindent In this section, we provide extensive experiments to verify the effectiveness of the proposed \textbf{ATP} based on both ResNet-101 and MiT-B5 backbones. Specifically, we reveal the improvement of proposed components on GTA5$\rightarrow$Cityscapes task by progressively adding each module into the system. As Table \ref{Tab:Abl} shows, we can observe that each component contributes to adaptation, achieving the final performance of 52.6\% mIoU and 64.0\% mIoU on the source data-free scenario. 


    \begin{table}[!tbp]
    \centering
    \tabcolsep=5pt
    \caption{Ablation study of each component on GTA5$\rightarrow$ Cityscapes domain adaptation task based on different ResNet-101 and MiT-B5 backbones.}
    \label{Tab:Abl}
    \begin{tabular}{ccccccccc}
    \toprule[1pt]
     Model     & $\mathcal{L}_{cel}$  & $\mathcal{L}_{div}$ & $\mathcal{L}_{div}^{*}$ & $\mathcal{L}_{ppl}$ & $\mathcal{L}_{npl}$ & $\mathcal{L}_{ssl}$ & mIoU \\
    \midrule[0.7pt]
    \multicolumn{8}{c}{\textbf{ResNet-101}} \\
    \midrule[0.7pt]
    Baseline &  & & & & & & 39.3  \\
    (a)         & \Checkmark & & & & & & 43.6 \\
    (b)         & \Checkmark & \Checkmark & & & & & 45.0 \\
    (c)         & \Checkmark &  & \Checkmark & & & & 43.9 \\ 
    (d)         & \Checkmark & \Checkmark &  & \Checkmark &  & & 47.7  \\ 
    (e)         & \Checkmark & \Checkmark & & \Checkmark & \Checkmark & & 49.8 \\ 
    (f)         & \Checkmark & \Checkmark & & \Checkmark & \Checkmark & \Checkmark & \textbf{52.6} \\ 
    \midrule[0.7pt]
    \multicolumn{8}{c}{\textbf{MiT-B5}} \\
    \midrule[0.7pt]
    Baseline &  & & & & & & 44.5  \\
    (g)         & \Checkmark & & & & & & 51.8 \\
    (h)         & \Checkmark & \Checkmark & & & & & 54.0 \\
    (i)         & \Checkmark &  & \Checkmark & & & & 53.2 \\ 
    (j)         & \Checkmark & \Checkmark &  & \Checkmark &  & & 60.3  \\ 
    (k)         & \Checkmark & \Checkmark & & \Checkmark & \Checkmark & & 62.4 \\ 
    (l)         & \Checkmark & \Checkmark & & \Checkmark & \Checkmark & \Checkmark & \textbf{64.0} \\ 
    \bottomrule[1.0pt]
    \end{tabular}
    \end{table}

    \textbf{Influence of curriculum feature alignment.} From the models (a) and (b) in Table \ref{Tab:Abl}, we can observe that our curriculum-style entropy loss ($\mathcal{L}_{cel}$) and weighted diversity loss ($\mathcal{L}_{div}$) can significantly improve adaptation performance. Specifically, model (a) with $\mathcal{L}_{cel}$ achieves a 43.6\% mIoU result, which surpasses the baseline model by 4.3 points. It also significantly outperforms the previous entropy minimization method \cite{UncertaintyReducing} reported mIoU result as 19.8\%.
    Furthermore, model (b) with $\mathcal{L}_{div}$ boosts the performance improvement with 1.4 mIoU, revealing that diversity enforcing is important for our setting. 
    $\mathcal{L}_{div}^{*}$ denotes diversity loss provided by \cite{SHOT} with the mean output embedding. Compared model (b) with (c), we see that both diversity-promoting terms are beneficial for our task, and the proposed weighted one ($\mathcal{L}_{div}$) works better than a non-weighted one ($\mathcal{L}_{div}^{*}$). 

    \textbf{Influence of curriculum complementary self-training.} To verify the effectiveness of the proposed complementary self-training method, we ablate the positive pseudo labeling $\mathcal{L}_{ppl}$ and negative pseudo labeling $\mathcal{L}_{npl}$, respectively. The results show that model (d) trained from the only positive pseudo labeling provides 2.7 point gains compared to model (b). It is reasonable that the positive pseudo-labels provide useful supervised information for the target data. On the other hand, adding the negative pseudo labeling (model (e)) fancily offers 2.1 point improvement. This result demonstrates that negative pseudo labels can provide complementary information, which is ignored by the traditional positive pseudo labeling methods. Both the positive and negative pseudo-labeling are helpful in learning decision boundaries.

    \textbf{Influence of information propagation.} To testify the effectiveness of the proposed information propagation strategy, we report the result of model (f) in Table \ref{Tab:Abl}, which provides a significant improvement with 2.8 mIoU compared to model (e). It demonstrates our information propagation technique is effective in reducing intra-domain discrepancy.

    \begin{table}[!tbp]
        \centering
        \tabcolsep=18pt
        \caption{Domain adaptation results of GTA5$\rightarrow$Cityscapes task based on VGG-16 network.}
        \begin{tabular}{lcc}
        \toprule[1.0pt]
         Method  & SF & mIoU  \\
        \midrule
        Baseline             & - & 27.1 \\
        ADVENT \cite{ADVENT} & \XSolidBrush & 35.6 \\
        AdaptSegNet \cite{AdaptSeg} & \XSolidBrush & 35.0 \\
        CLAN \cite{CLAN} & \XSolidBrush & 36.6 \\
        CyCADA \cite{CyCADA} & \XSolidBrush & 35.4 \\
        PatchDA \cite{PatchDA} & \XSolidBrush & 37.5 \\
        SSF-DAN \cite{SSF-DAN} & \XSolidBrush & 37.7 \\
        \midrule
        \textbf{ATP (Ours)} & \Checkmark &  \textbf{39.8} \\
        \bottomrule[1.0pt]
        \end{tabular}
        \label{Tab:VGG}
    \end{table}

    \begin{table}[!tbp]
    \centering
    \tabcolsep=10pt
    \caption{Results of different weights $\lambda$ for weighted diversity loss in Eq. (\ref{Eq:div}) on GTA5$\rightarrow$Ciyscapes task.}
    \label{Tab:weight}
    \begin{tabular}{cccccc}
    \toprule[1.0pt]
        $\lambda$ & 1 & 2 & 3 & 4 & 5  \\
    \midrule[0.7pt]
         mIoU & 44.0 & 44.2 & \bf{45.0} & 44.1 &  44.0\\
    \bottomrule[1.0pt]
    \end{tabular}
    \end{table}

    \begin{figure*}[!tbp]
        \centering
        \includegraphics[width=0.96\linewidth]{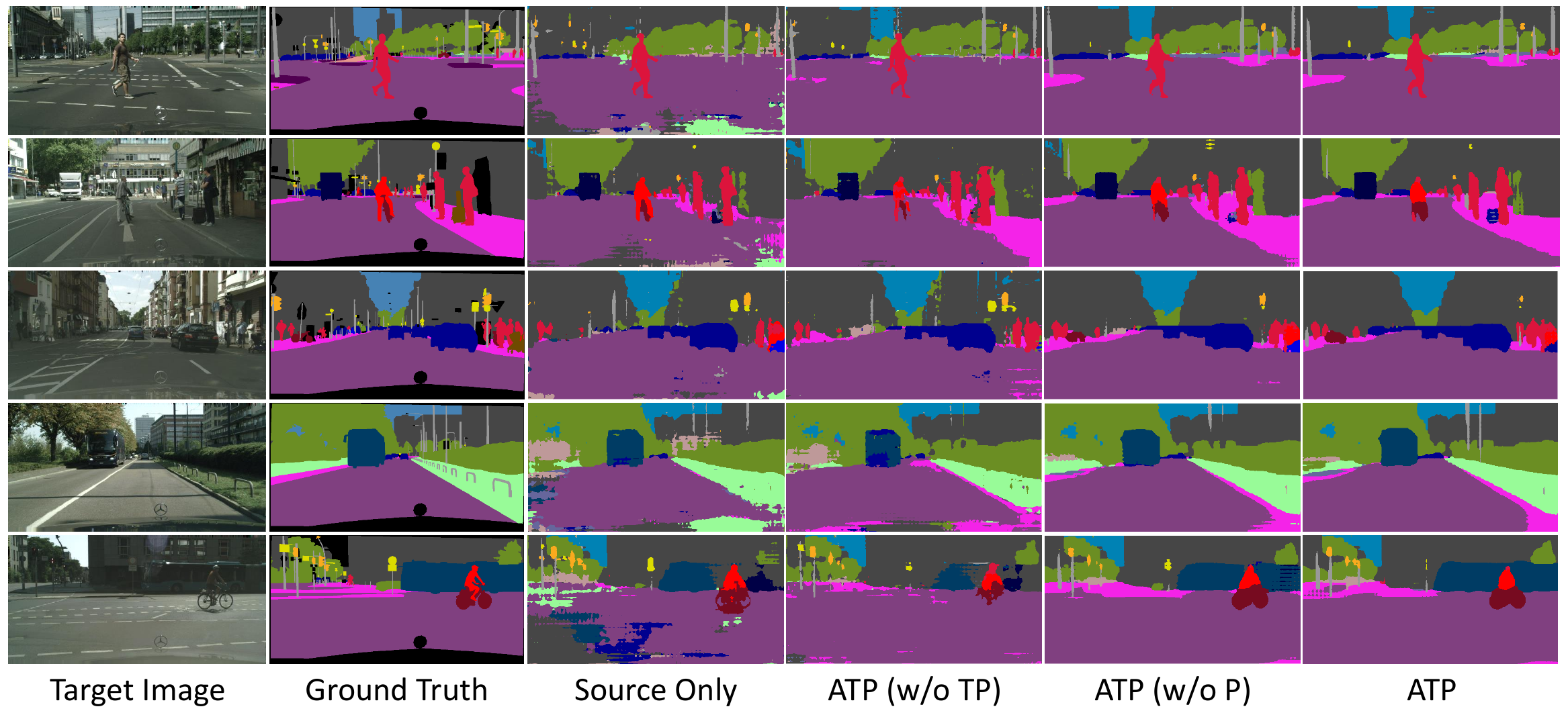}
        \caption{Visualization for predicted segmentation masks on the GTA5$\rightarrow$Cityscapes task.}
        \label{Fig:Vis}
        \vspace{-0.3cm}
    \end{figure*}
    
\begin{figure}[!tbp]
    \centering
    \includegraphics[width=0.98\linewidth,trim=280 140 280 150,clip]{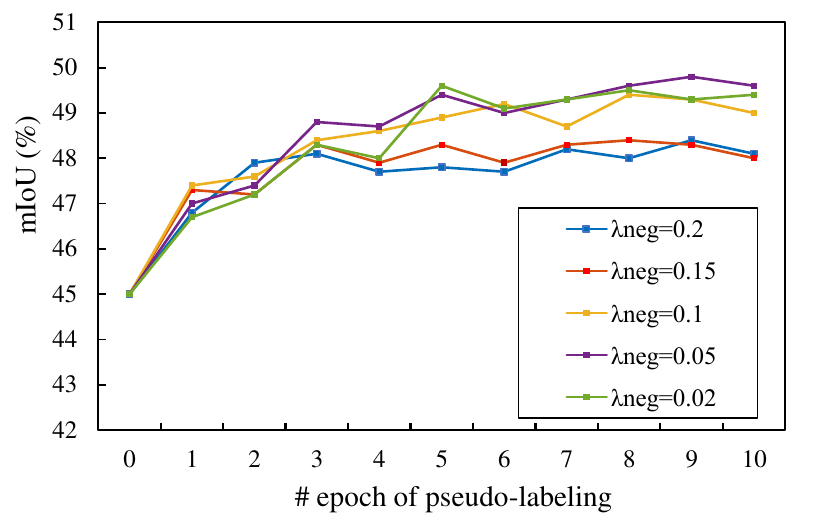}
    \caption{Robustness to the negative threshold $\lambda_{neg}$ for the negative pseudo assignment method.}
    \label{Fig:ParamLambda}
\end{figure}

    \textbf{Performance on different backbones.} To demonstrate the robustness of our method, we also apply different backbones to conduct the semantic segmentation network. In this section, we utilize VGG-16 \cite{VGG} networks as a segmentation feature extractor, and the final adaptation performance is shown in Table \ref{Tab:VGG}. We compare existing source data-dependent domain adaptation works that report results based on VGG-16 networks. We can observe that our method surpasses existing conventional source data-dependent domain adaptation approaches by achieving 39.8\% mIoU, which reveals the proposed ATP is robust for different backbone networks.

    It should be noticed that a similar tendency is observed when we applied a Transformer-based backbone, MiT-B5. Specifically, each of the proposed components performs stable improvement for the source data-free adaptation.

\subsection{Parameter Sensitivity Analysis}

\begin{figure}[!tbp]
    \centering
    \includegraphics[width=0.98\linewidth,trim=280 140 280 150,clip]{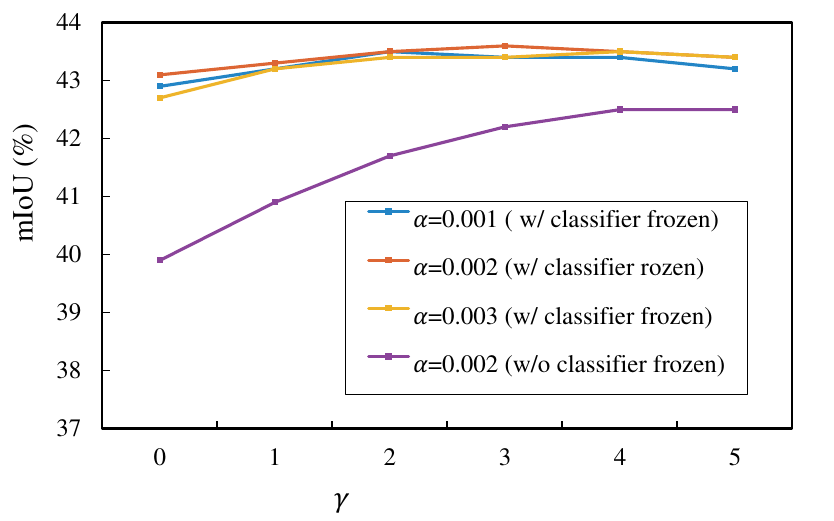}
    \caption{Analysis for the hyper-parameters $\alpha$ and $\gamma$ of the proposed adaptive entropy loss.}
    \label{Fig:HyperParam}
\end{figure}

    \noindent Our framework contains two new hyper-parameters $\alpha$ and $\gamma$ in Eq.~(\ref{Eq:fel}). To analyze the influence, we conduct experiments on the GTA5$\rightarrow$Cityscapes task using ResNet-101 backbone. We first freeze $\gamma$ and choose $\alpha=0.002$ from the range of [0.001, 0.002, 0.003], and then the value of $\gamma$ is selected as 3 from the range of [0, 1, 2, 3, 4, 5]. The results are shown in Figure \ref{Fig:HyperParam}. Although all experiments set $\alpha=0.002$ and $\gamma=3$ since it peaks, we observe that the proposed curriculum-style entropy loss is relatively robust to these parameters. Besides, we also optimize the model without freezing the source classifier. The performance drops significantly compared to those with a frozen classifier, which indicates that the hypothesis transfer with a frozen classifier is essential.
    
    Figure \ref{Fig:ParamLambda} shows results when varying the assignment threshold $\lambda_{neg}$ related to negative pseudo labels in Eq.~(\ref{Eq:NPL}). It shows that the proposed negative pseudo labeling method is relatively robust to this hyper-parameter. Moreover, the proposed method quickly converges after several training epochs. We also observe that $\lambda_{neg} < 0.1$ obtains a satisfactory performance compared to those when $\lambda_{neg}>0.1$. The reason is that more noise labels are introduced when the threshold increases, as our negative pseudo labeling method focuses on predictions with lower confidence scores.

    More hyper-parameters introduced in this paper are analyzed in this subsection. We provide the risk of target supervision leak in the complementary self-training stage. 
    For example, in Table \ref{Tab:weight}, we analyze the weight $\lambda$ for weighted diversity loss in Eq. (\ref{Eq:div}), which controls the influence of weighted mean output $\hat{p}_{x_t}^{(h,w,c)}$.

\subsection{Visualization}

    \noindent In Figure \ref{Fig:Vis}, we visualize predicted segmentation masks obtained from our method and compare the results to the masks predicted by the ``Source-Only" network. To show the contribution of each stage, we add each component progressively and provide qualitative results of ``ATP (w/o TP)", ``ATP (w/o P)", and ``ATP", respectively. It is evident that even without source data, our method achieves reliable results with fewer spurious areas. Furthermore, compared to the ``Source Only" model, each stage contributes significantly in terms of pixel-level accuracy.

\section{Conclusion}

    \noindent This paper presents a novel source data-free approach for domain adaptive semantic segmentation called \textbf{ATP}. In the absence of source data, \textbf{ATP} focuses on handling model adaptation via the proposed curriculum-style self-training, including curriculum feature alignment to encourage confident predictions and curriculum complementary self-training using negative pseudo labeling and positive pseudo labeling. We also provide a standard post-processing for domain adaptation by propagating semantic consistency information to reduce the intra-domain discrepancy.
    Extensive experiments and ablation studies are conducted to validate the effectiveness of the proposed method. \textbf{ATP} achieves new state-of-the-art results and performs comparably to source data-dependent adaptation methods. In addition, a simplified variant of ATP performs well on the black-box source model scenario. 
    
\section*{Acknowledgement}
    \noindent The authors would like to thank the reviewers and the associate editor for their valuable comments. 
    This work was supported in part by the National Key R\&D Program of China (No. 2022ZD0116500), the National Natural Science Foundation of China (No. U21B2042), and in part by the 2035 Innovation Program of CAS, and the InnoHK program.

\ifCLASSOPTIONcaptionsoff
  \newpage
\fi



%
{\small
\bibliographystyle{ieee}
\bibliography{mybib}
}
%

\begin{IEEEbiography}[{\includegraphics[width=1in,height=1.25in,clip,keepaspectratio]{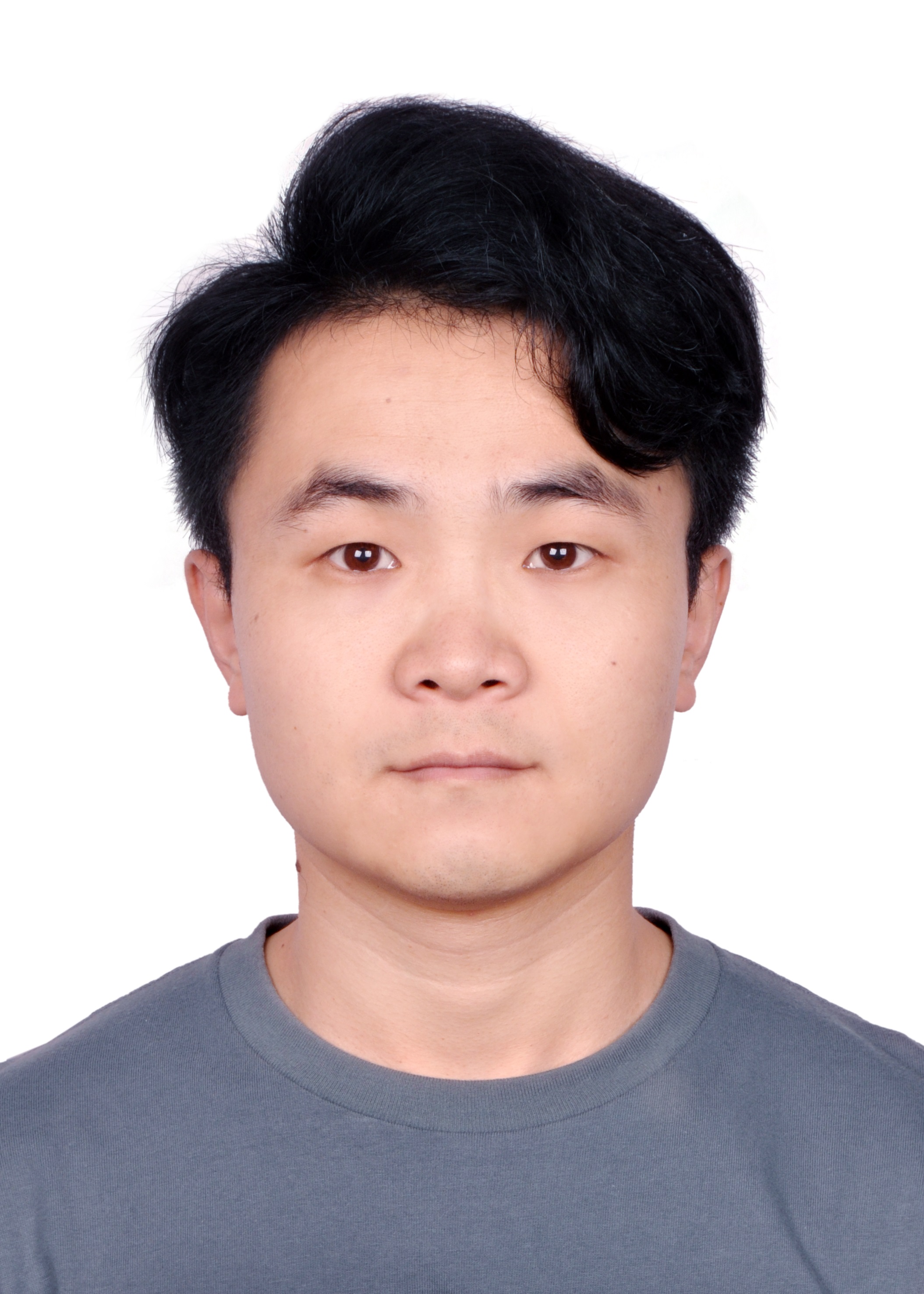}}]{\textbf{Yuxi Wang}}
received the B.E. degree from North Eastern University, China, in 2016, and the PhD degree from University of Chinese Academy of Sciences (UCAS), Institute of Automation, Chinese Academy of Sciences (CASIA), in January 2022. He is now an assistant professor in the Centre for Artificial Intelligence and Robotics (CAIR), Hong Kong Institute of Science \& Innovation, Chinese Academy of Science (HKISI-CAS). His research interests include transfer learning, domain adaptation, multimodal large language model, and computer vision.
\end{IEEEbiography}

\begin{IEEEbiography}[{\includegraphics[width=1in,height=1.25in,clip,keepaspectratio]{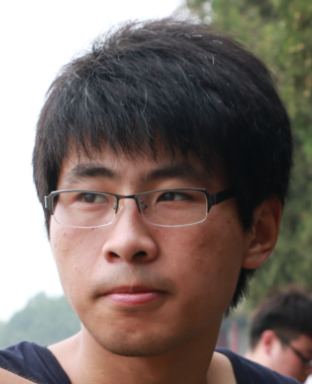}}]{Jian Liang}
received the B.E. degree in Electronic Information and Technology from Xi'an Jiaotong University and Ph.D. degree in Pattern Recognition and Intelligent Systems from NLPR, CASIA in July 2013, and January 2019, respectively.
He was a research fellow at National University of Singapore from June 2019 to April 2021. Now he joins State Key Laboratory of Multimodal Artificial Intelligence Systems as an associate professor. His research interests focus on transfer learning, pattern recognition, and computer vision.
\end{IEEEbiography}

\begin{IEEEbiography}[{\includegraphics[width=1in,height=1.25in,clip,keepaspectratio]{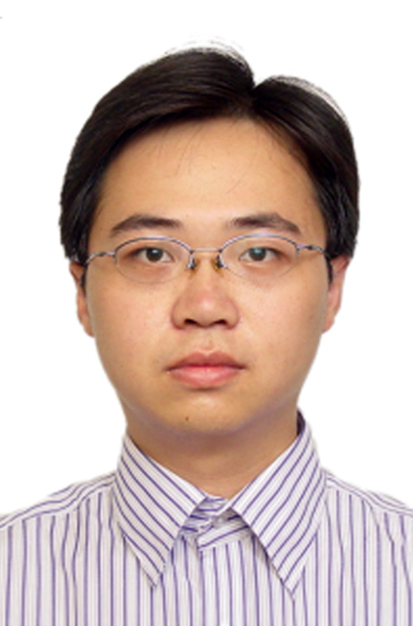}}]{\textbf{Zhaoxiang Zhang}} received his bachelor’s degree from the Department of Electronic Science and Technology in the University of Science and Technology of China (USTC) in 2004. After that, he was a Ph.D. candidate under the supervision of Professor Tieniu Tan in the National Laboratory of Pattern Recognition, Institute of Automation, Chinese Academy of Sciences, where he received his Ph.D. degree in 2009. In October 2009, he joined the School of Computer Science and Engineering, Beihang University, as an Assistant Professor (2009-2011), an Associate Professor (2012-2015) and the Vise-Director of the Department of Computer application technology (2014-2015). In July 2015, he returned to the Institute of Automation, Chinese Academy of Sciences. He is now a full Professor in the New Laboratory of Pattern Recognition (NLPR) and the State Key Laboratory of Multimodal Artificial Intelligence Systems.
His research interests include Computer Vision, Pattern Recognition, and Machine Learning. Recently, he specifically focuses on deep learning models, biologically-inspired visual computing and human-like learning, and their applications on human analysis and scene understanding.
He has published more than 200 papers in international journals and conferences, including reputable international journals such as JMLR, IEEE TIP, IEEE TNN, IEEE TCSVT, IEEE TIFS and top level international conferences like CVPR, ICCV, NIPS, ECCV, AAAI, IJCAI and ACM MM.
He is serving or has served as the Associate Editor of IEEE T-CSVT, Patten Recognition, Neurocomputing, and Frontiers of Computer Science. He has served as the Area Chair, Senior PC of international conferences like CVPR, NeurIPS, ICML, AAAI, IJCAI and ACM MM. He is a Senior Member of IEEE, a Distinguished Member of CCF, and a Distinguished member of CAAI.
\end{IEEEbiography}

\vfill



\end{document}